\documentclass{article}
\usepackage[utf8]{inputenc}

\usepackage{graphicx}
\usepackage{amssymb}
\usepackage{amsthm}
\usepackage{url}
\usepackage{amsmath}
\usepackage{subfigure}
\usepackage[affil-it]{authblk}
\usepackage[margin=2.5cm]{geometry}
\usepackage{lscape}
\usepackage[final]{changes}

\definechangesauthor[name=Alex, color=purple]{AB}
\definechangesauthor[name=Rene, color=blue]{RL}

\graphicspath{{figs/}}

\title{\replaced[id=AB]{Mem-fractive}{Memristive} Properties of Mushrooms}
\author[1,*]{Alexander E. Beasley\footnote{Corresponding author: Alexander Beasley, alex.beasley@uwe.ac.uk}}
\author[3]{Mohammed-Salah Abdelouahab}
\author[2]{Ren\'{e} Lozi}
\author[1]{Anna L. Powell}
\author[1]{Andrew Adamatzky}
\affil[1]{Unconventional Computing Laboratory, UWE, Bristol, UK}
\affil[2]{Universit\'{e} C\^{o}te d’Azur, CNRS, LJAD, Nice, France}
\affil[3]{Laboratory of Mathematics and their interactions, University Centre Abdelhafid Boussouf, Mila 43000, Algeria}
\date{}

\begin{document}

\maketitle

\begin{abstract}
\noindent

Memristors close the loop for I-V characteristics of the traditional, passive, semi-conductor devices. Originally proposed in 1971, 
the hunt for the memristor has been going ever since. The key feature of a memristor is that its current resistance is a function of its previous resistance and the current passed through it. As such, the behaviour of the device is influenced by changing the way in which potential is applied across it. Ultimately, information can be encoded on memristors. Biological substrates have already been shown to exhibit some memristive properties.
\added[id=AB]{ By extension, the mem-capacitor and mem-inductor have been proposed. Such devices change either their capacitive or inductive properties a function of the previous voltage, similar to memristors. A device that exhibits combinations of memristors, mem-capacitors and mem-inductors is termed a mem-fractive device.}
However, many \replaced[id=AB]{passive memory}{memristive} devices are yet to be found. Here we show that the fruit bodies of grey oyster fungi \emph{Pleurotus ostreatus} exhibit \added[id=AB]{encouraging} behaviour \added[id=AB]{in the field of organic memory devices}. This paper presents the I-V characteristics of the mushrooms. By examination of the conducted current for a given voltage applied as a function of the previous voltage, it is shown that the mushroom  \replaced[id=AB]{exhibits the properties of a mem-fractor}{memristor}. Our results demonstrate that nature continues to provide specimens that hold these unique and valuable electrical characteristics and which  have the potential to advance the field of hybrid electronic systems.   

\vspace{5mm}


\end{abstract}

\section{Introduction}

Originally proposed by Chua in 1971~\cite{paper:memristor}, the memristor poses a fourth basic circuit element, whose characteristics differ from that of R, L and C elements. Memristance has been seen in nano-scale devices where electronic and ionic transport are coupled under an external bias voltage~\cite{paper:missing_memristor}. Strukov~\emph{et al.} posit that the hysteric I-V characteristics observed in thin-film, two-terminal devices can be understood as memristive.  However, this is observed behaviour of devices that already have other, large signal behaviours.

Finding a true memristor is by no means an easy task. Nevertheless, a number of studies have turned to nature to provide the answer, with varying success. 
Memristive properties of organic polymers were discovered well before the `official' discovery of the memristor  was announced in ~\cite{paper:missing_memristor}. \added[id=RL]{The first examples of memristors could go back to the singing arc, invented by Duddell in 1900, and originally used in wireless telegraphy before the invention of the triode~\cite{chaoscnns}.} Memristive properties of organic polymers have been studied since 2005~\cite{erokhin2005hybrid} in experiments with hybrid electronic devices based on polyaniline-polyethylenoxide junction~\cite{erokhin2005hybrid}.
Memristive properties of living creatures and their organs and fluids have been demonstrated in 
skin~\cite{martinsen2010memristance},
blood~\cite{kosta2011human},
plants~\cite{volkov2014memristors} (including fruits~\cite{paper:apples_memristor}), 
slime mould~\cite{gale2015slime},
tubulin microtubules~\cite{del2019two,chiolerio2020resistance}.

This paper presents a study of the I-V characteristics of the fruit bodies of the grey oyster fungi \emph{Pleurotus ostreatus}. Why fungi? Previously we recorded extracellular electrical potential of oyster's fruit bodies, basidiocarps~\cite{adamatzky2018spiking} and found that the fungi generate action potential like impulses of electrical potential. The impulses can propagate as isolated events, or in trains of similar impulses. Further, we demonstrated, albeit in numerical modelling, that fungi can be used as computing devices, where information is represented by spikes of electrical activity, a computation is implemented in a mycelium network and an interface is realised via fruit bodies~\cite{adamatzky2018towards}. A computation with fungi  might not be useful \emph{per se}, because the speed of spike propagation is substantially lower than the clock speed in conventional computers. However, the fungal computation becomes practically feasible when embedded in a slow developing spatial process, e.g. growing architecture structures. Thus, in ~\cite{adamatzky2019fungal} we discussed how to: produce adaptive building constructions by developing structural
substrate using live fungal mycelium, functionalising the substrate with nanoparticles and polymers to make mycelium-based electronics, implementing sensorial fusion and decision making in the fungal electronics.

Why we are looking for mem\replaced[id=AB]{-fractive}{ristor} properties of fungi? \added[id=AB]{Mem-fractors~\cite{Abdelouahab2014memfractance} have combinations of properties exhibited by memristors, mem-capacitors and mem-inductors.} A memristor is a material implication~\cite{borghetti2010memristive,kvatinsky2013memristor} and can, therefore, can be used for constructing other logical circuits, statefull logic operations~\cite{borghetti2010memristive}, logic operations in passive crossbar arrays of memristors~\cite{linn2012beyond}, memory aided logic circuits~\cite{kvatinsky2014magic}, self-programmable logic circuits~\cite{borghetti2009hybrid}, and, indeed, memory devices~\cite{ho2009nonvolatile}. If strands of fungal mycelium in a culture substrate and the fruit bodies show some mem\replaced[id=AB]{-fractive}{ristor} properties then we can implement a large variety of memory and computing devices embedded directly into architectural building materials made from the fungal substrates~\cite{adamatzky2019fungal}.

The rest of this paper is organised as follows. Section~\ref{sec:experimentation} details the experimental set up used to examine the I-V characteristics of fruit bodies. Section~\ref{sec:results} presents the results from the experimentation, with further discussion of voltage spiking provided in section~\ref{sec:spiking}. Mathematical modelling of the mem-fractive behaviour of the Grey Oyster mushrooms is given in section~\ref{sec:model}. A discussion of the results is given in section~\ref{sec:discussions} and finally conclusions are given in section~\ref{sec:results}. 

\section{Experimental Set Up}
\label{sec:experimentation}

\begin{figure}[!tbp]
    \centering
\subfigure[]{\includegraphics[width=0.49\textwidth]{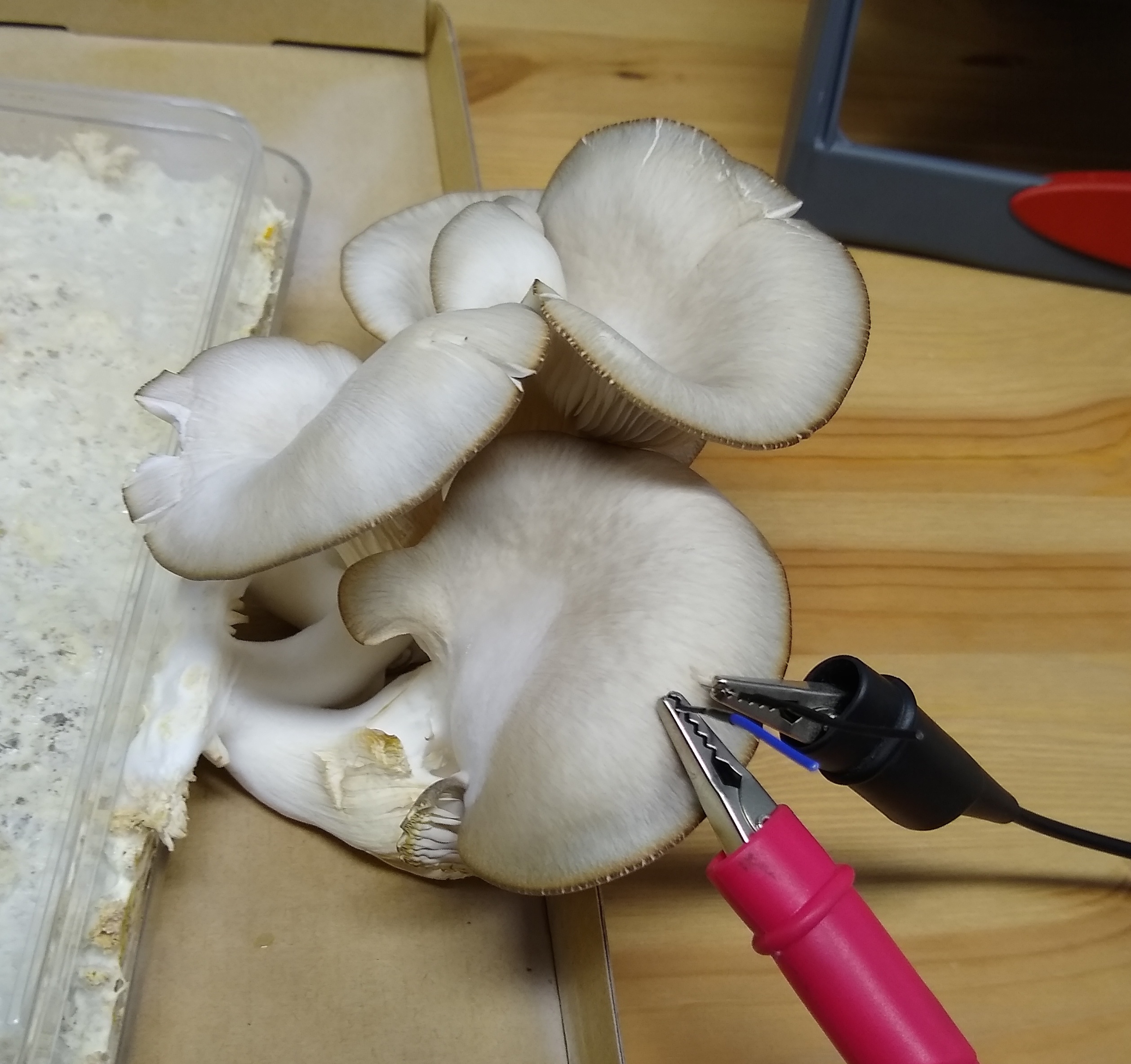}\label{fig:captocap}}
\subfigure[]{\includegraphics[width=0.49\textwidth]{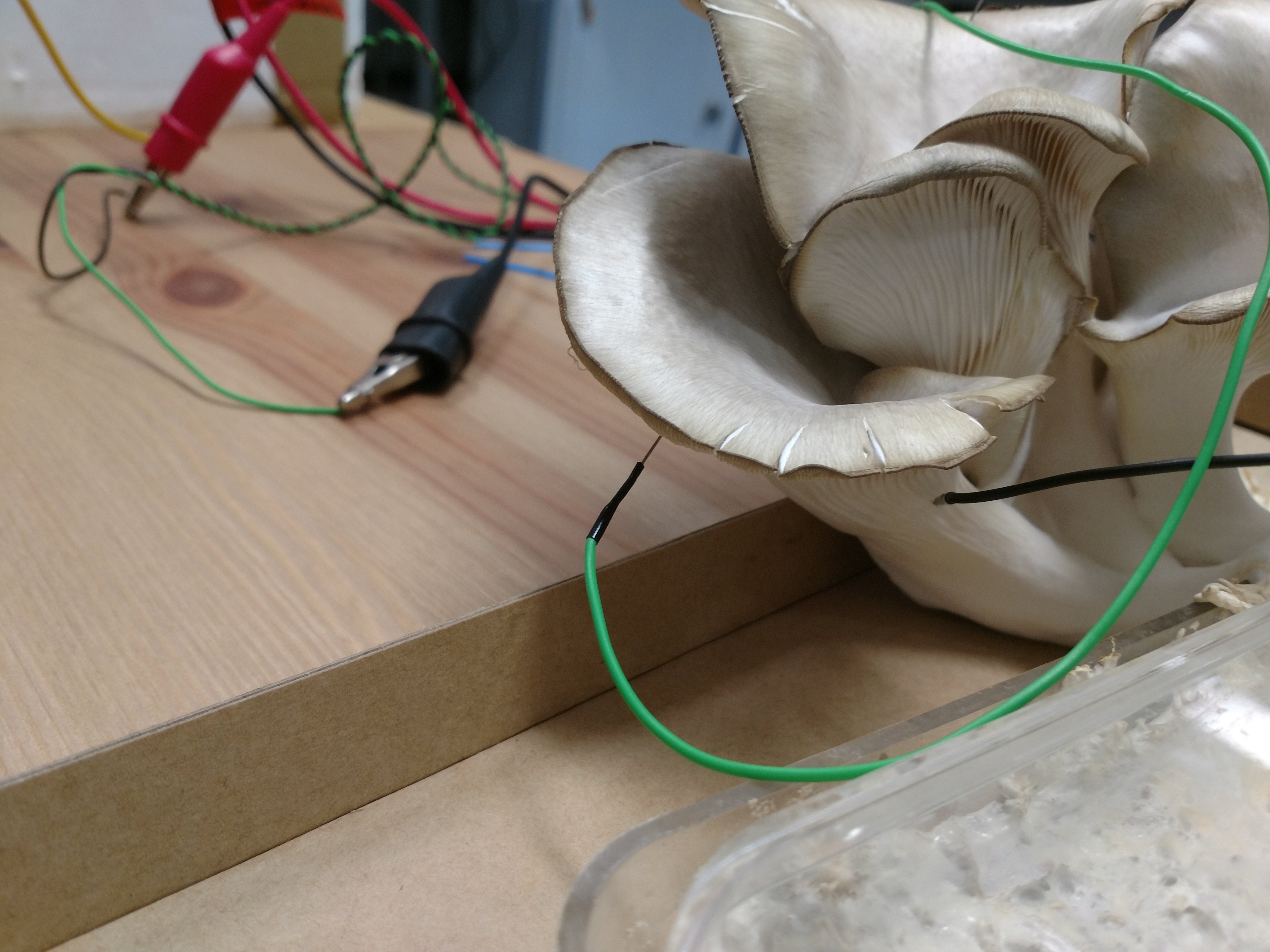}\label{fig:captostem}}
    \caption{Positions of electrodes in fruit bodies. (a)~Electrodes inserted 10~mm apart in the fruit body cap. (b)~One electrode is inserted in the cap with the other in the  stem. }
   \label{fig:fruitbodiesrecording}
\end{figure}

We used  grey oyster fungi \emph{Pleurotus ostreatus} (Ann Miller's Speciality Mushrooms Ltd, UK)  cultivated on wood shavings. The iridium-coated stainless steel sub-dermal needles with twisted cables (Spes Medica SRL, Italy)  were inserted in fruit bodies (Fig.~\ref{fig:fruitbodiesrecording}) of grey oyster fungi using two different arrangements: 10~mm apart in the cap of the fungi (cap-to-cap), Fig.~\ref{fig:captocap}, and translocation zones (cap-to-stem), Fig.~\ref{fig:captostem}. I-V sweeps were performed on the fungi samples with Keithley Source Measure Unit (SMU) 2450 (Keithley Instruments, USA) under the following conditions: [-500~mV to 500~mV, -1~V to 1~V] with the samples in ambient lab light (965~Lux). Varying the step size of the voltage sweep allowed testing the I-V characteristics of the subject at different frequencies. Electrodes were arranged in two different methods: both electrodes approximately 10~mm apart in the cap of the fruit body (Fig.~\ref{fig:captocap}); and one electrode placed in the cap with the other electrode placed in the stem (Fig.~\ref{fig:captostem}). The voltage ranges are limited so as not to cause the electrolysis of water. Each condition was repeated at least six times over the samples. Voltage sweeps were performed in both directions (cyclic voltammetry) and plots of the I-V characteristics were produced. 

MATLAB was used to analyse the frequency and distribution of spiking behaviour observed in the I-V sweeps of the fruiting bodies under test (Sect.~\ref{sec:spiking}). All histogram plots are binned according to the voltage interval set for the Kiethley SMU.\par

\section{Results}
\label{sec:results}

\subsection{I-V characterisation}

\begin{figure}[!tbp]
    \centering
    \includegraphics[width=0.8\textwidth]{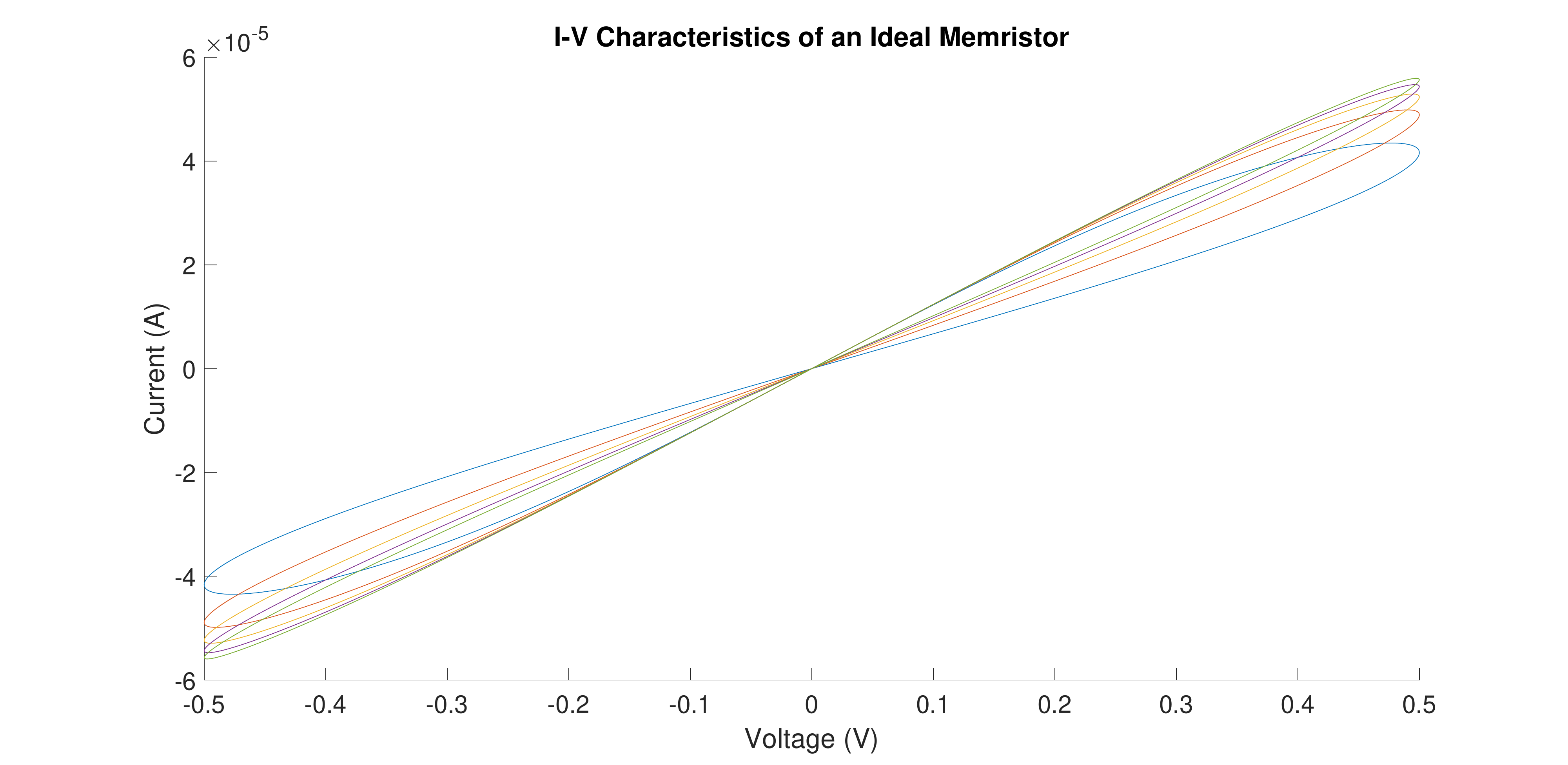}
    \caption{I-V Characteristics from a model of an ideal memristor~\cite{web:matlab_memristor_model}.}
    \label{fig:ideal}
\end{figure}

Fruit body samples  are shown to exhibit memristive properties when subject to a voltage sweep. The ideal memristor model (Fig.~\ref{fig:ideal}) is shown to display `lobes' on the I-V characterisation sweeps, indicating that the current resistance is a function of the previous resistance --- hence a memristor has memory. For the purposes of analysis, graphs are referred to by their quadrants, starting with quadrant one as the top right and being number anti-clockwise. \par

\begin{figure}[!tbp]
    \centering
\subfigure[]{       \includegraphics[width=0.7\textwidth]{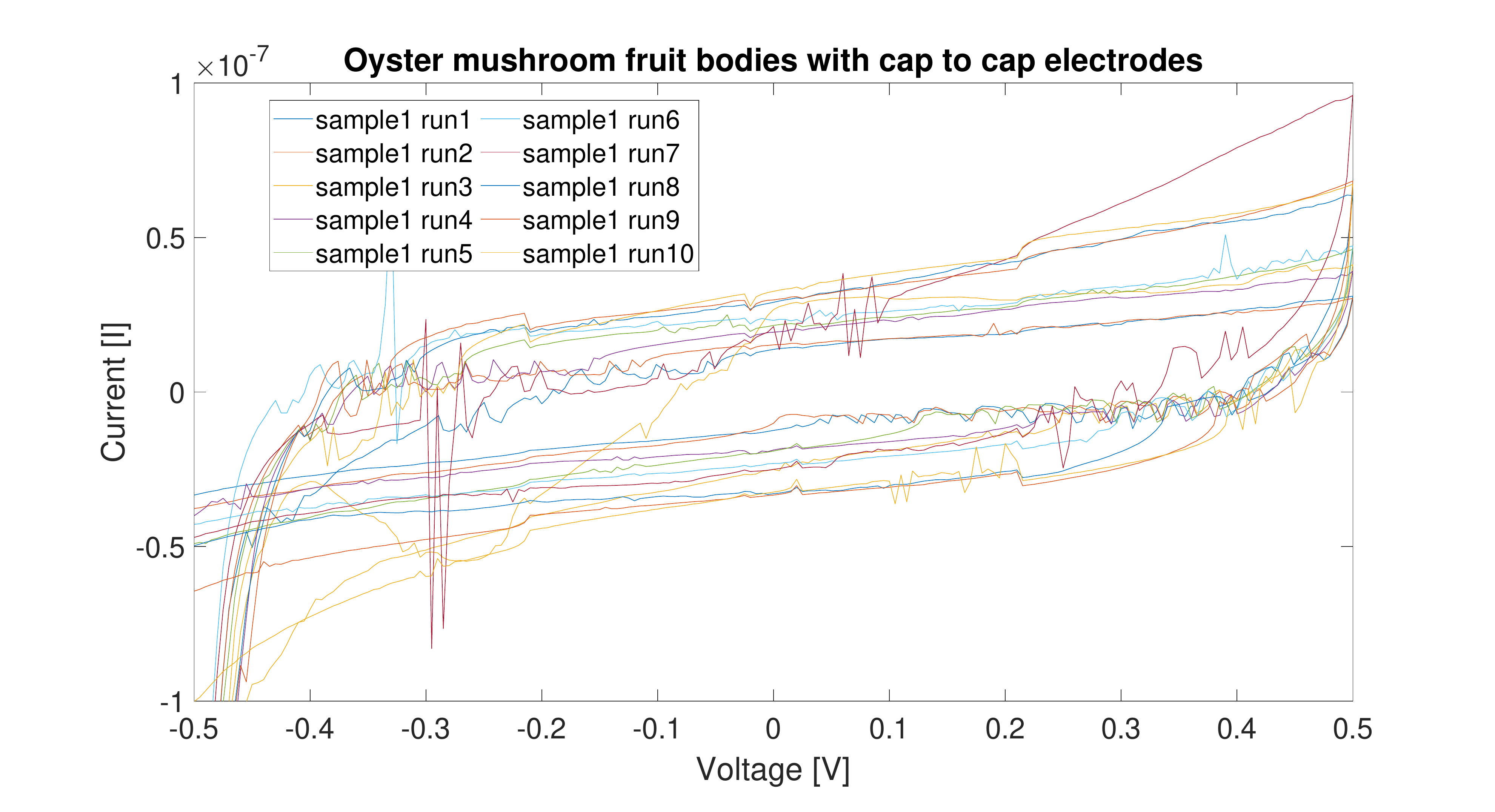}}
\subfigure[]{
        \includegraphics[width=0.7\textwidth]{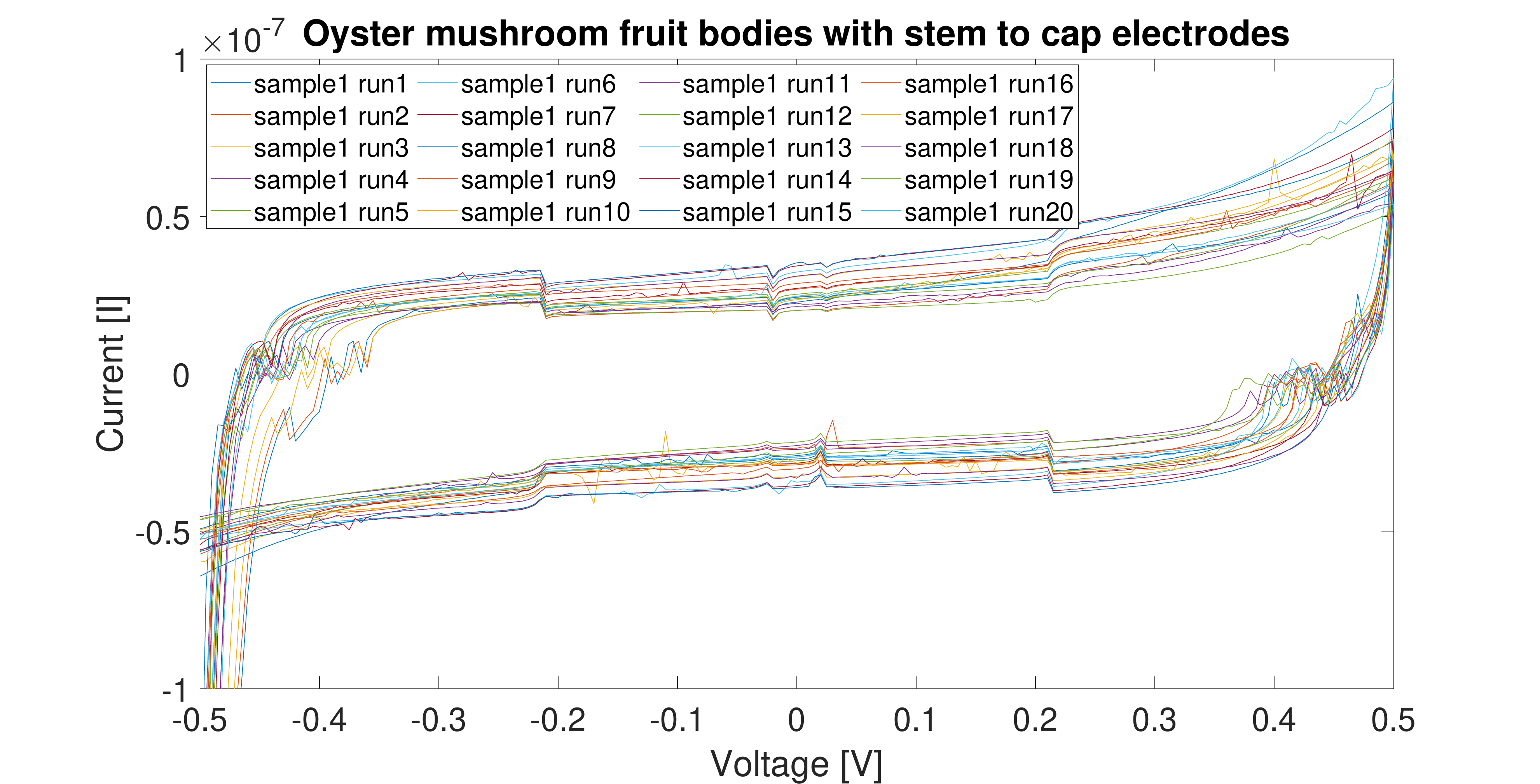}}
    \caption{Raw data from cyclic voltammetry performed over -0.5~V to 0.5~V. (a)~Cap-to-cap electrode placement. (b)~Stem-to-cap electrode placement.}
    \label{fig:oysterfruit1V}
\end{figure}

\begin{figure}[!tbp]
    \centering
\subfigure[]{
        \includegraphics[width=0.7\textwidth]{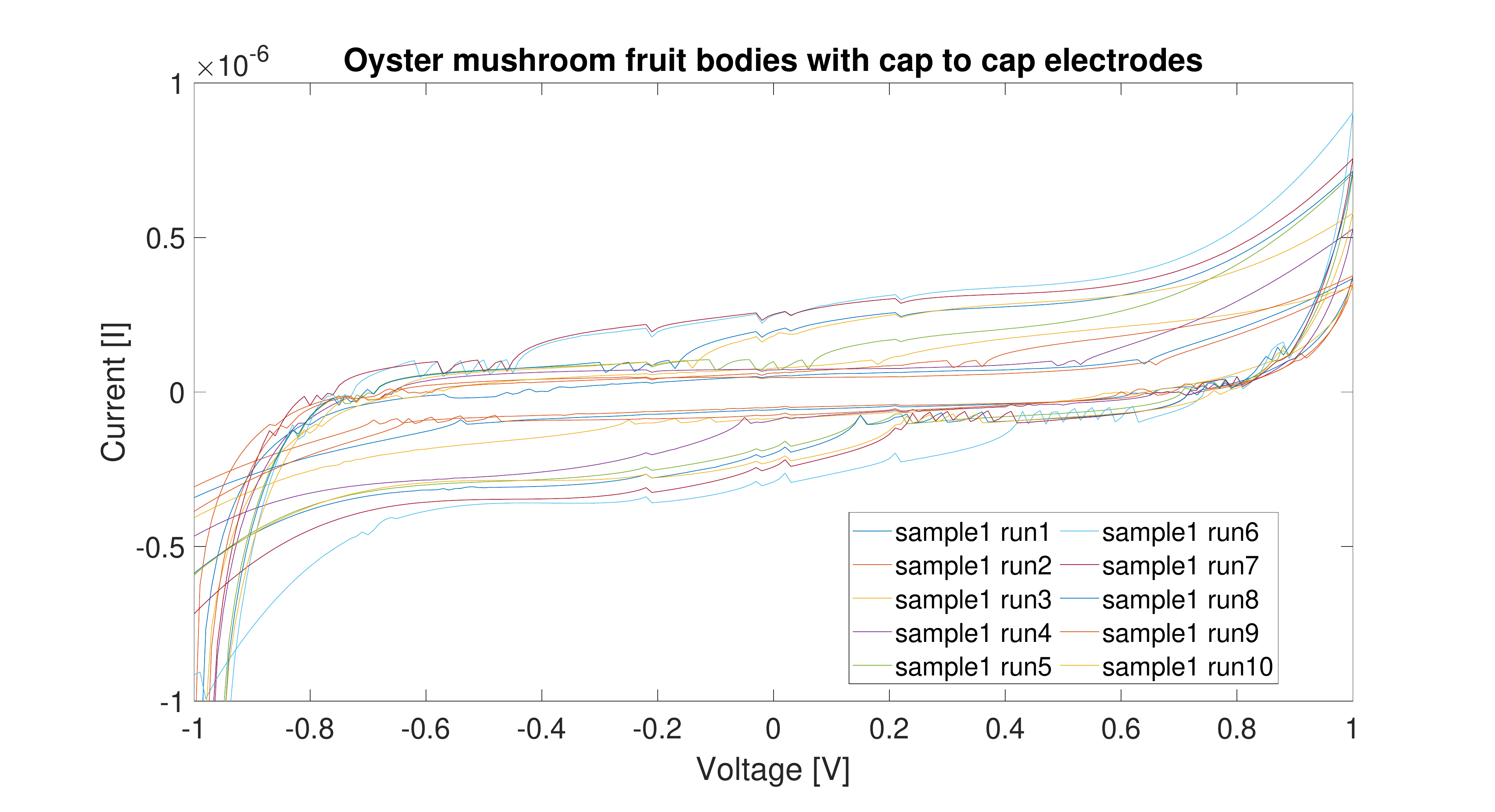}}
\subfigure[]{
        \includegraphics[width=0.7\textwidth]{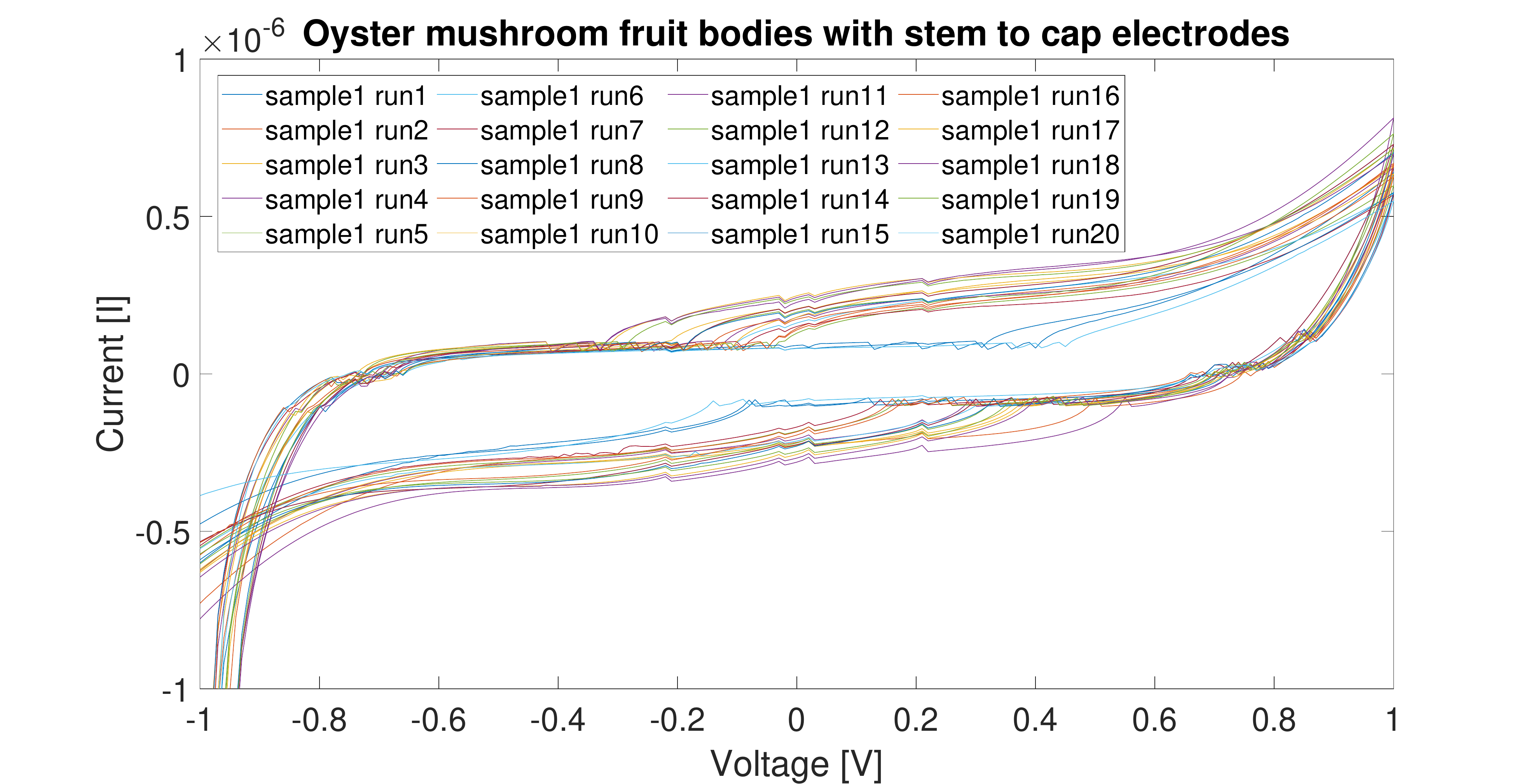}}
    \caption{Raw data from cyclic voltammetry performed over -1~V to 1~V. (a)~Cap-to-cap electrode placement. (b)~Stem-to-cap electrode placement.}
    \label{fig:oysterfruit2V}
\end{figure}

The ideal memristor model has a crossing point at 0V, where theoretically no current flows. From Figs.~\ref{fig:oysterfruit1V} and \ref{fig:oysterfruit2V}, it can be seen that when 0~V is applied by the source meter, a reading of a nominally small voltage and current is performed. The living membrane is capable of generating potential across the electrodes, and hence a small current is observed. \added[id=AB]{Mem-capacitors produce similar curves to that in Fig.~\ref{fig:ideal}, when plotting charge ($q$) against voltage ($v$)~\cite{memcapacitor2015Yin}. Additionally, mem-inductors produce similar plots for current ($i$) against flux ($\varphi$).}\par

While the sample under test is subjected to a positive voltage (quadrant 1), it can be seen there is nominally a positive current flow. Higher voltages result in a larger current flow. For an increasing voltage sweep there is a larger current flow for the corresponding voltage during a negative sweep.\par

Similarly, in quadrant 3 where there is a negative potential across the electrodes, the increasing voltage sweep yields a current with smaller magnitude than the magnitude of the current on a negative voltage sweep.\par

Put simply, the fruit body has a resistance that is a function of the previous voltage conditions.\par 

\begin{figure}[!htb]
\centering
\subfigure[]{
        \includegraphics[width=0.7\textwidth]{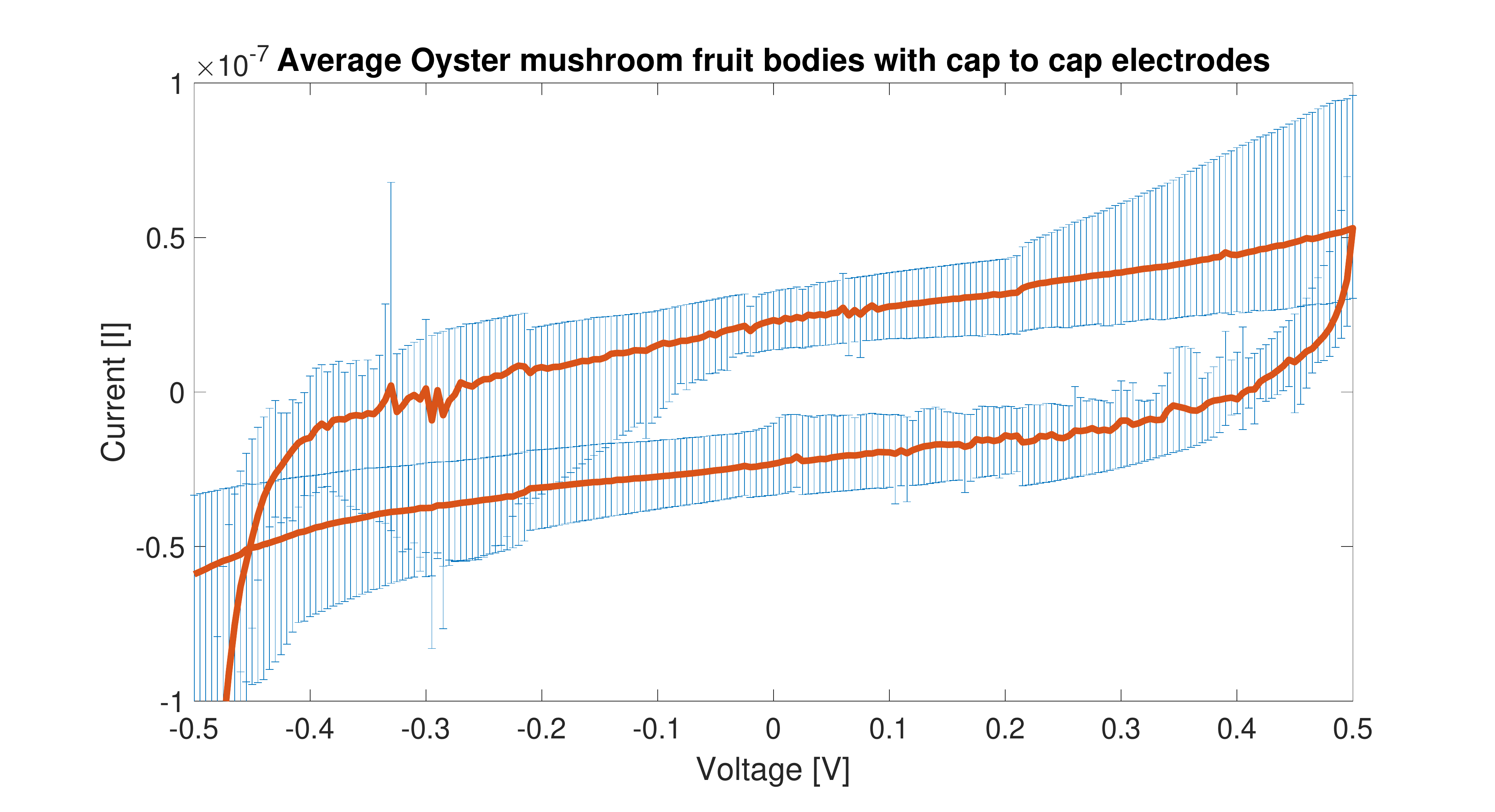}\label{fig:1V_average_cap_to_cap}}
\subfigure[]{
        \includegraphics[width=0.7\textwidth]{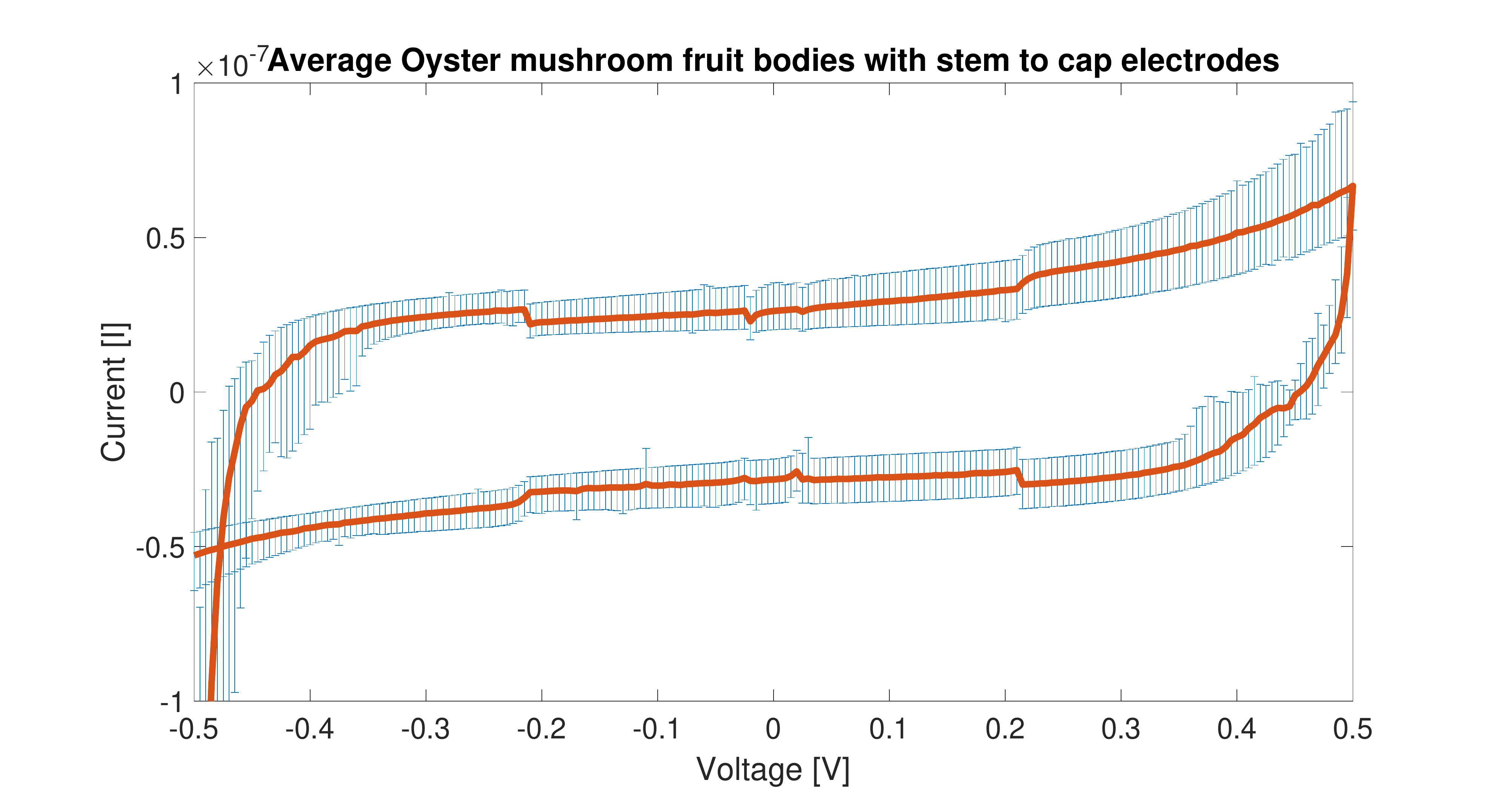}\label{fig:1V_average_stem_to_cap}}
    \caption{Average grey oyster fungi fruit bodies I-V characteristics for cyclic voltammetry of -0.5~V to 0.5~V. (a)~Cap-to-cap electrode placement. (b)~Stem-to-cap electrode placement.}
    \label{fig:1V_sweeps_oyster}
\end{figure}

\begin{figure}[!htb]
\centering
\subfigure[]{\includegraphics[width=0.7\textwidth]{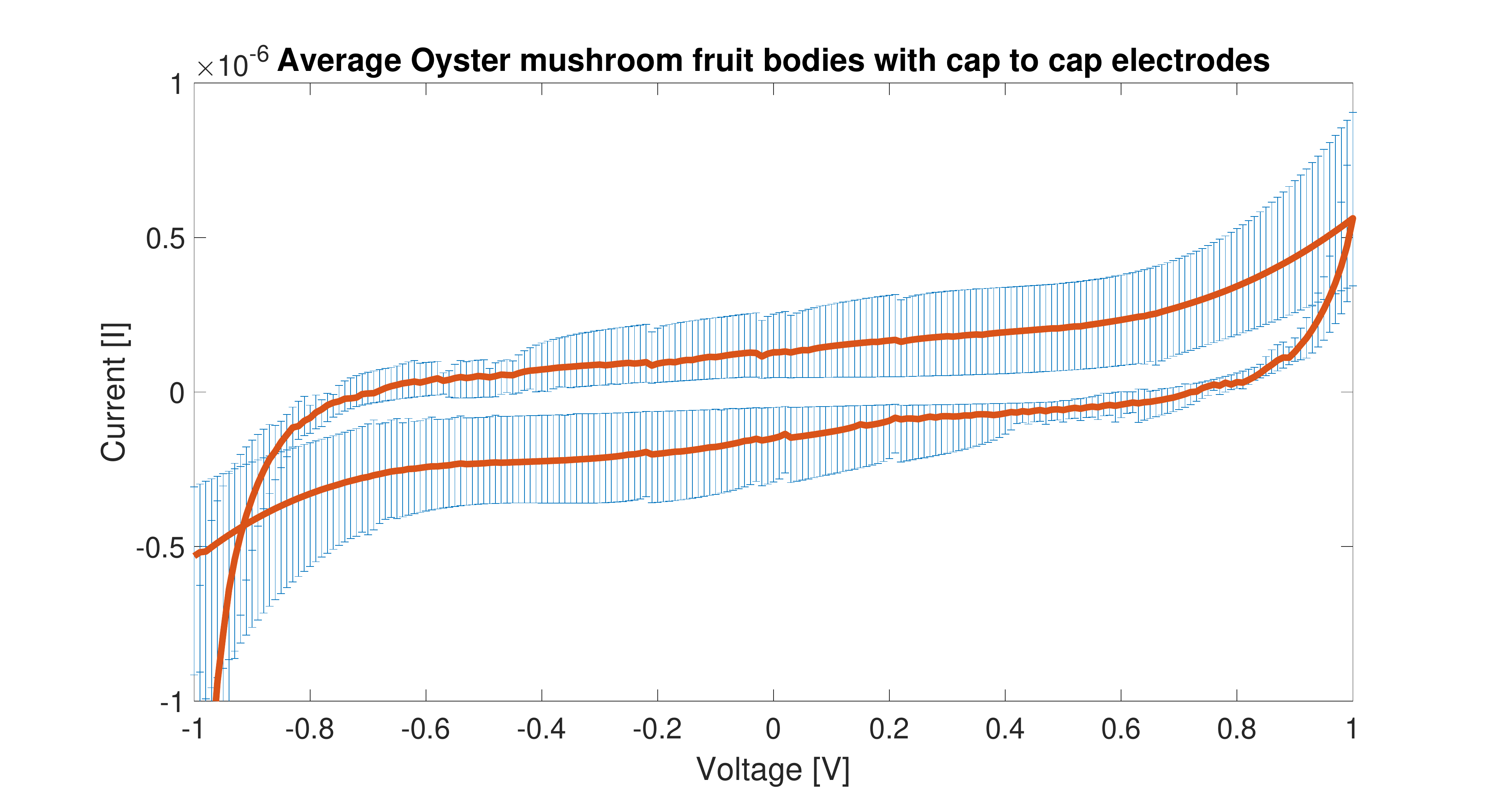}\label{fig:2V_average_cap_to_cap}}
\subfigure[]{       \includegraphics[width=0.7\textwidth]{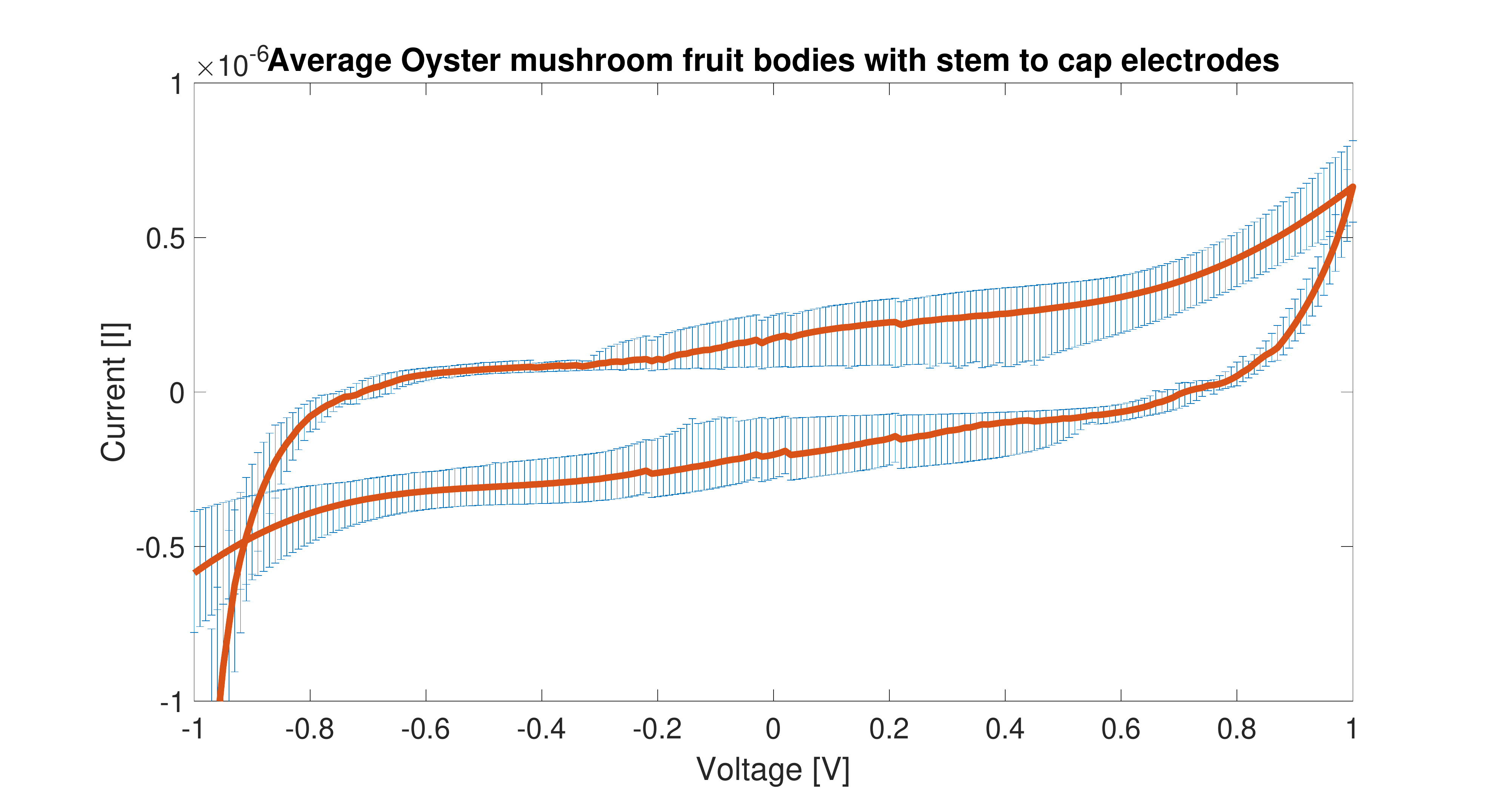}\label{fig:2V_average_stem_to_cap}}
    \caption{Average fruit bodies I-V characteristics for cyclic voltammetry of -1~V to 1~V. (a)~Cap-to-cap electrode placement. (b)~Stem-to-cap electrode placement.}
    \label{fig:2V_sweeps_oyster}
\end{figure}

By applying averaging to the performed tests, a clear picture is produced that demonstrates for a given set of conditions, a typical response shape can be expected (Figs.~\ref{fig:1V_sweeps_oyster} and \ref{fig:2V_sweeps_oyster}). The stem-to-cap placement of the electrodes in the fruit body yields a tighter range for the response (figures~\ref{fig:1V_average_stem_to_cap} and \ref{fig:2V_average_stem_to_cap}). This can be expected due to the arrangement of the transportation pathways, so-called translocation zone distinct from any vascular hyphae~\cite{schutte1956translocation,jennings1987translocation}, in the fruit body which run from the edge of the cap and down back through the stem to the root structure (mycelium). Cap-to-cap placement of the electrodes applies the potential across a number of the solutes translocation pathways and hence yields a wider range of results. However, for all results, it is observed that the positive phase of the cyclic voltammetry produces a different conduced current than the negative phase.\added[id=AB]{ The opening of the hysteresis curve around the zero, zero point suggests the fungus is not strictly a mem-ristor, instead it is also exhibiting mem-capacitor and mem-inductor effects. The build of charge in the device prevents the curve from closing completely to produce the classic mem-ristor pinching shape.}\par

\begin{figure}[!tbp]
    \centering
    \includegraphics[width=0.7\textwidth]{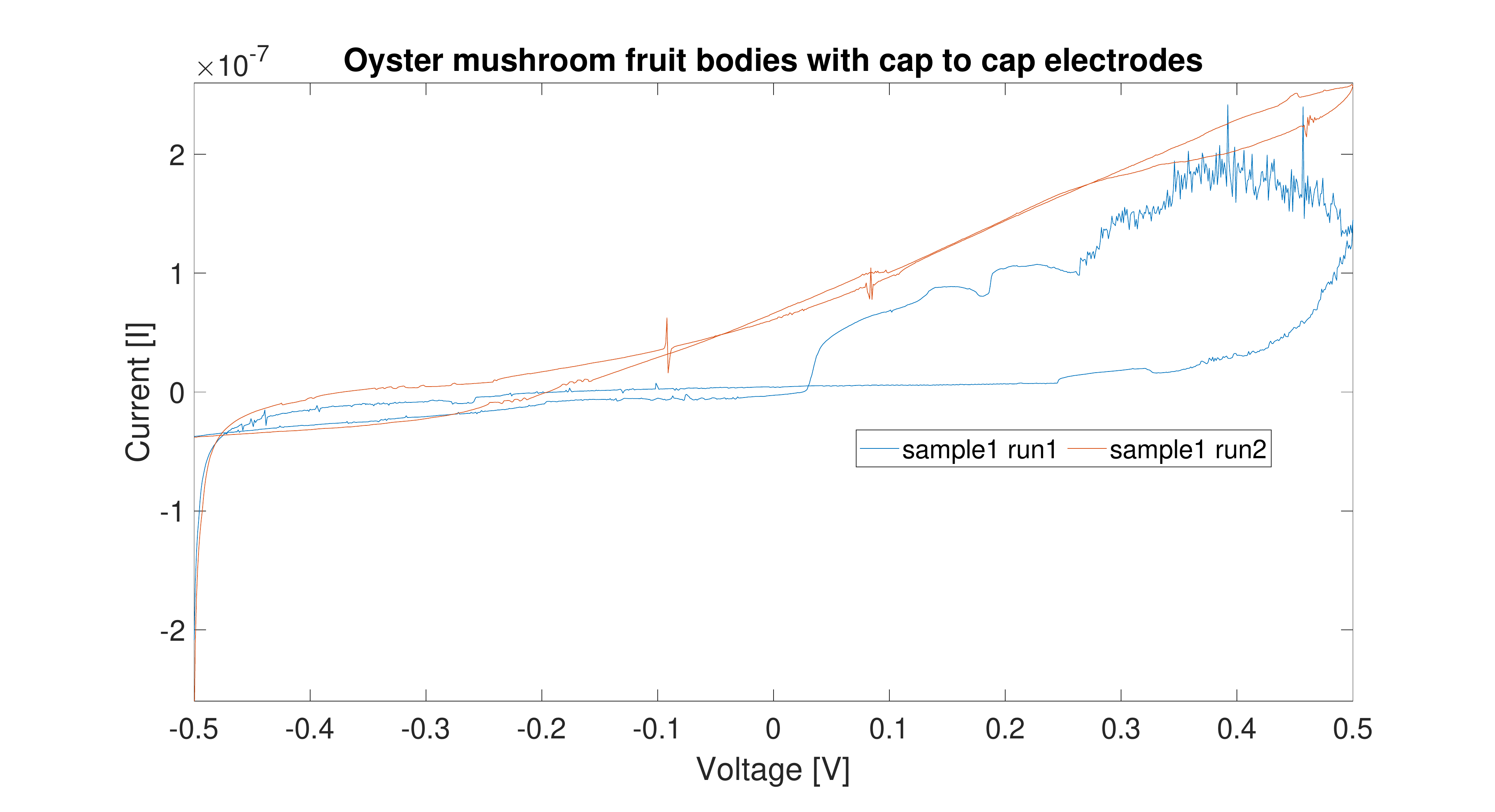}
    \caption{I-V Characteristics of fungi fruit bodies with the voltage step size set to 0.001~V.}
    \label{fig:slow}
\end{figure}

Reducing the step voltage step size (by ten fold) for the I-V characterisation is synonymous to reducing the frequency of the voltage sweep. Decreasing the sweep frequency of the voltage causes the chances of ``pinching'' in the I-V sweep to increase, as seen in quadrant 1 of figure~\ref{fig:slow}. \added[id=AB]{This further reinforces the presence of some mem-capacitor behaviour. Since the charging frequency of the fungus has now been reduced there is a greater amount of time for capacitively stored energy to dissipate, thus producing a more `resistive' plot with a pinch in the hysteresis.}\par

\subsection{Spiking}
\label{sec:spiking}

It is observed from Figs.~\ref{fig:oysterfruit1V} to \ref{fig:oysterfruit2V} that portions of the cyclic voltammetry result in oscillations in the conduced current, or spiking activity.  Oscillations occur most prominently on the positive phase of the cyclic voltammetry as the applied voltage approaches 0V and similarly on the negative phase, again as the applied voltage approaches 0~V. Current oscillations are typically in the order of nano-amps and persist for a greater number of cycles when the electrodes are arranged as a pair on the fruit body cap (between five and ten cycles) compared to the stem-to-cap arrangement (fewer than five repeats).\par

\begin{figure}[!tbp]
    \centering
    \includegraphics[width=\textwidth]{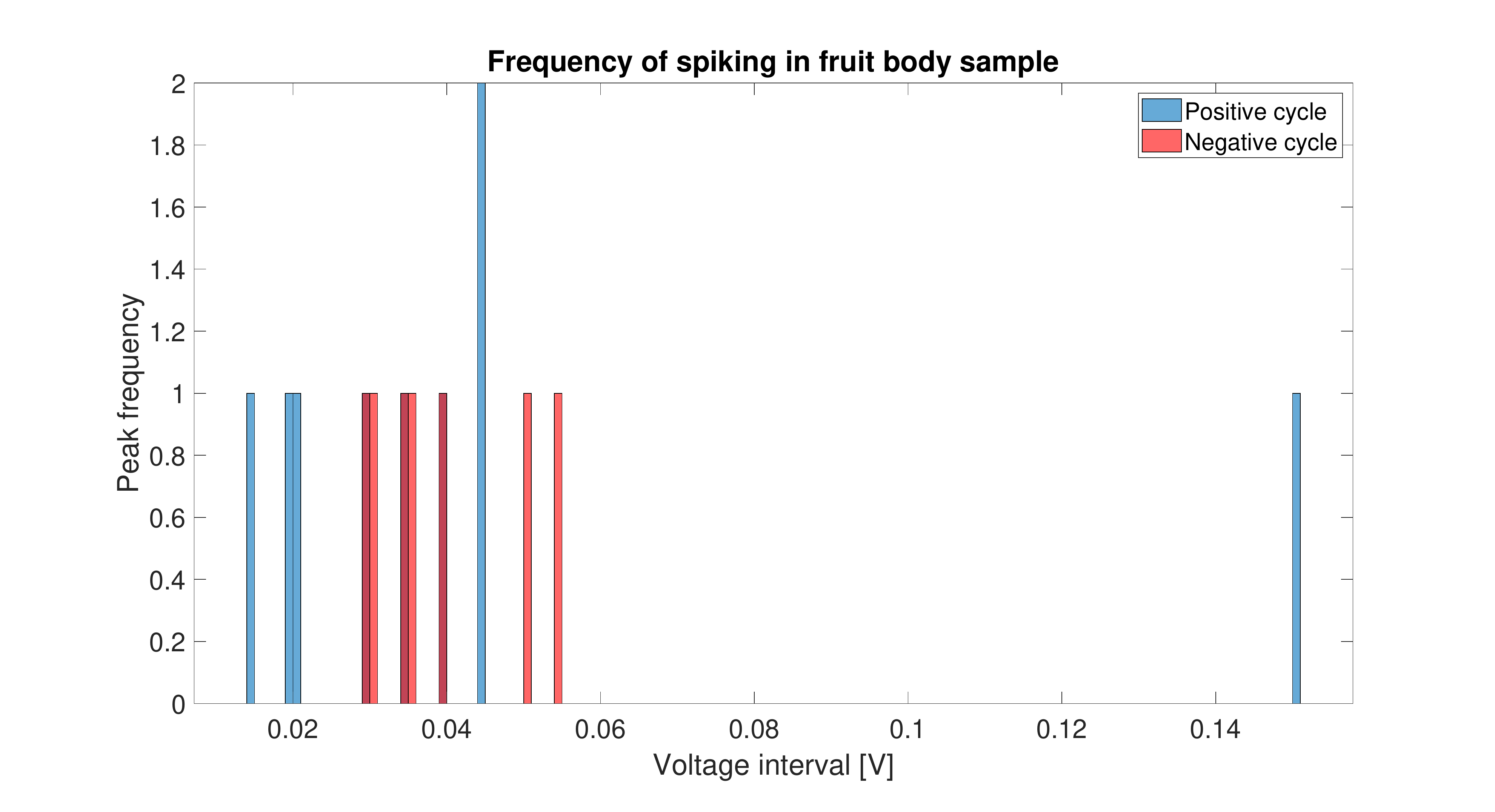}
    \caption{The voltage interval of spikes in the I-V characteristics of the fruit body for a single run.}
    \label{fig:captocap1Vpphist}
\end{figure}

\begin{figure}[!tbp]
    \subfigure[]{
        \includegraphics[width=0.49\textwidth]{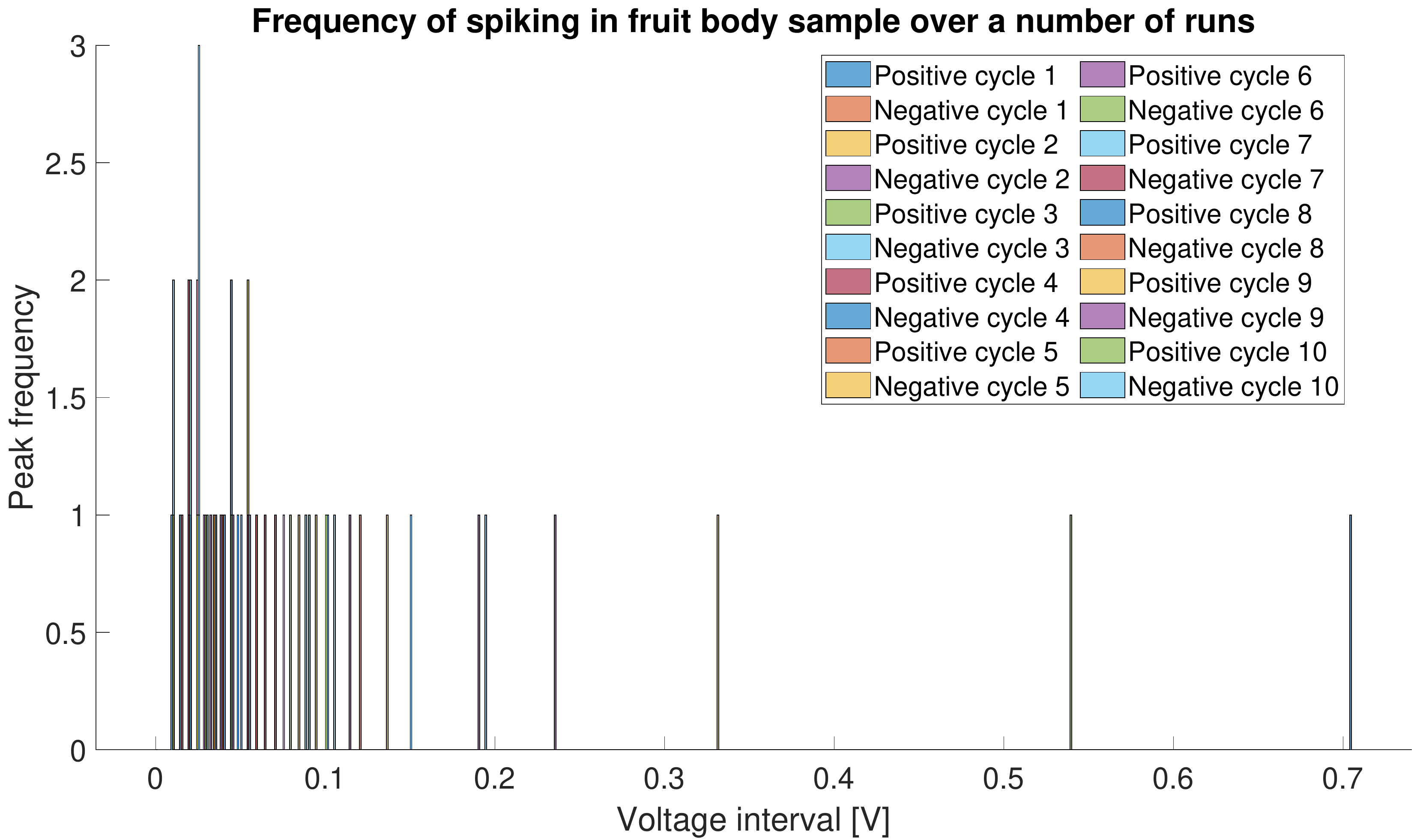}
        \label{fig:allspikes1Vppcaptocap}
        }
    \subfigure[]{\includegraphics[width=0.49\textwidth]{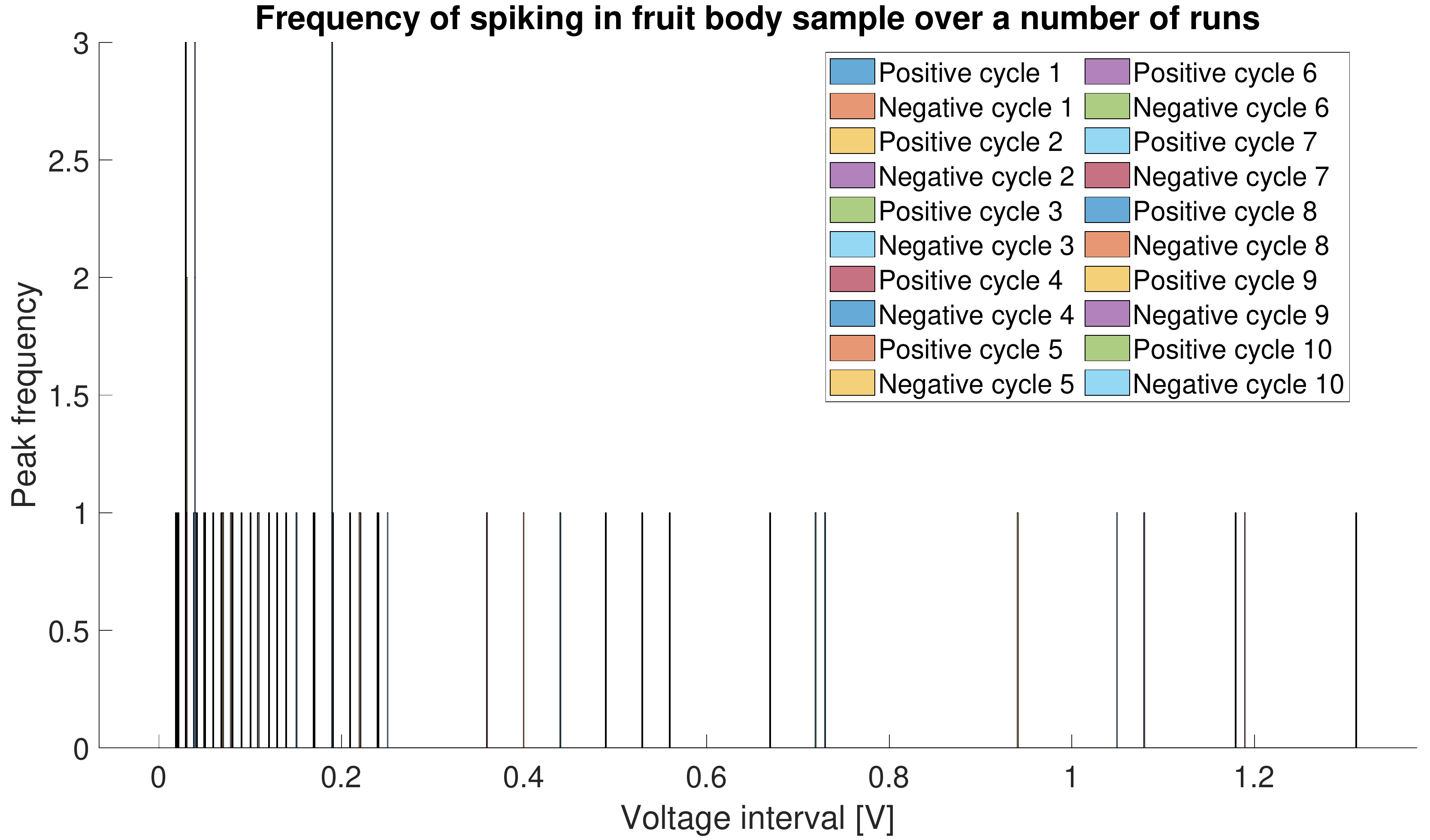}
    \label{fig:allspikes2Vppcaptocap}
    }
    \subfigure[]{\includegraphics[width=0.49\textwidth]{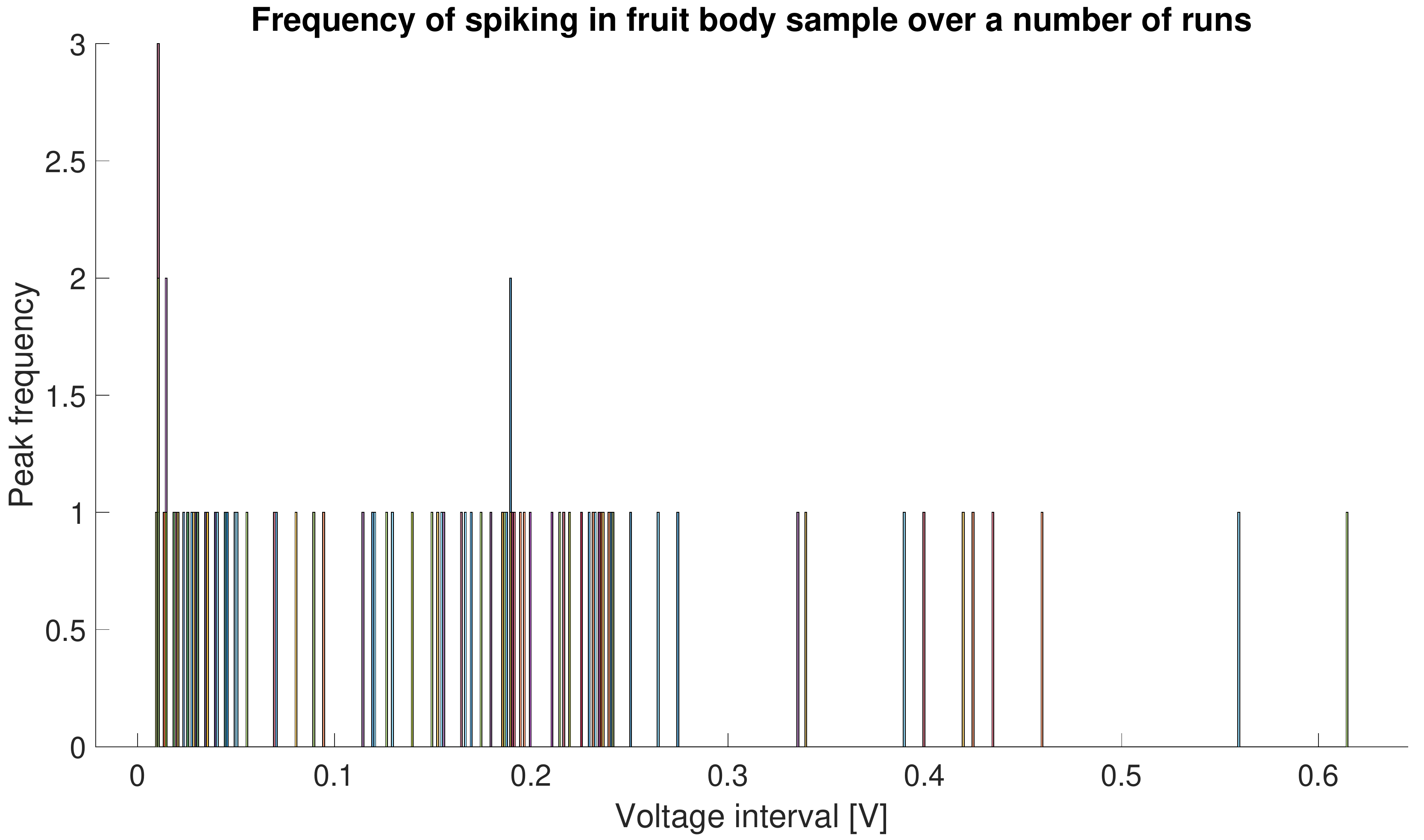}
    \label{fig:allspikes1Vppstemtocap}
    }
    \subfigure[]{\includegraphics[width=0.49\textwidth]{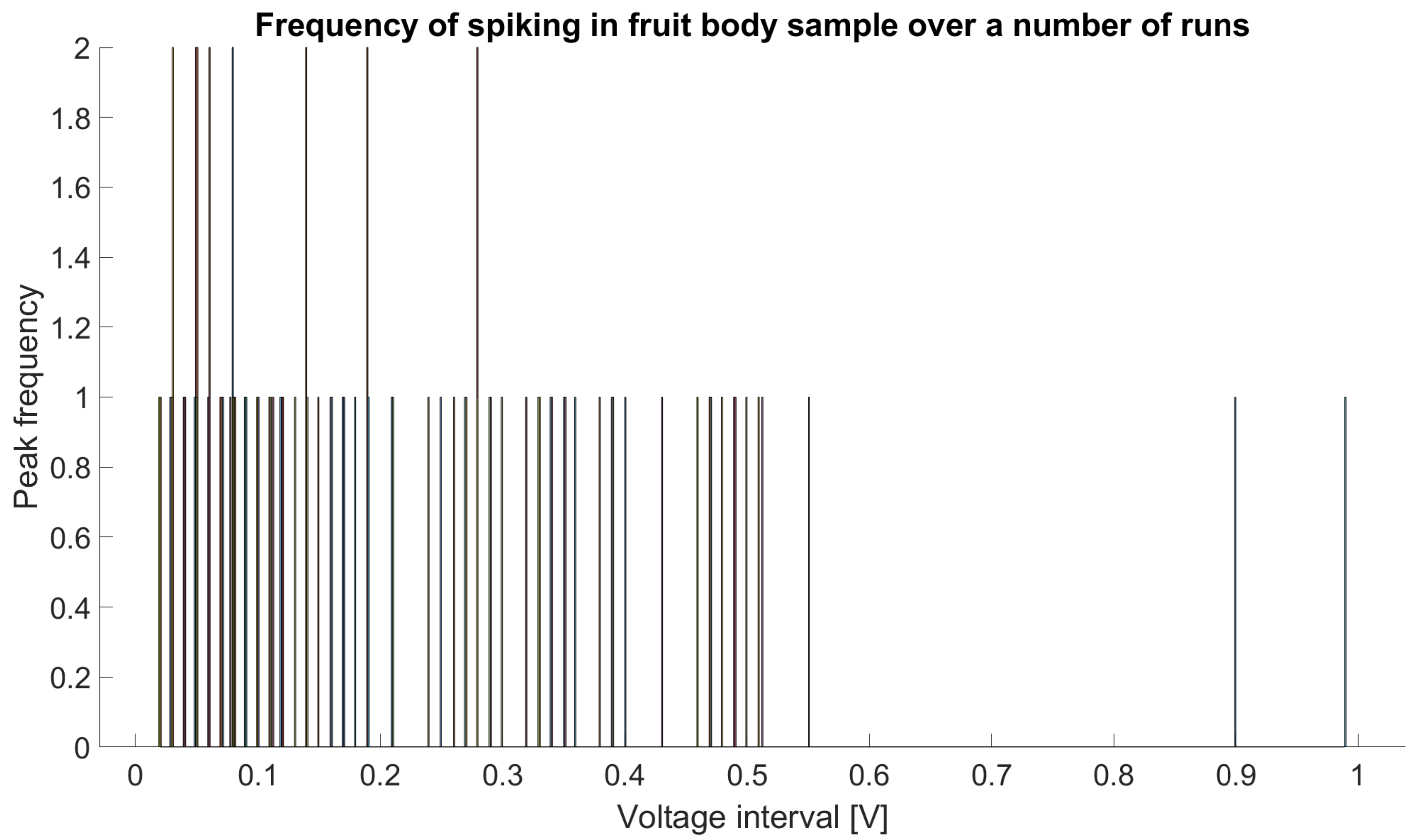}
    \label{fig:allspikes2Vppstemtocap}
    }
    \caption{Concatenations of all spiking data from all data runs for four different test conditions. (a) voltammetry over -0.5~V to 0.5~V, cap-to-cap electrode arrangement. (b) voltammetry over -1~V to 1~V, cap-to-cap electrode arrangement. (c) voltammetry over -0.5~V to 0.5~V, stem-to-cap electrode arrangement. (d) voltammetry over -1~V to 1~V, stem-to-cap electrode arrangement. Legends ommited on (c) and (d) for clarity.}
    \label{fig:all_hists}
\end{figure}

Figure~\ref{fig:captocap1Vpphist} demonstrates the spiking frequency of a single repeat of the cyclic voltammetry performed between -0.5~V and 0.5~V with the electrodes in a cap-to-cap arrangement. It is shown in the figure that the voltage interval between spikes in an oscillation period are less than 0.06~V. Figure~\ref{fig:all_hists} concatenates the data for all repeats of the cyclic voltammetry performed under four different conditions. It is clearly shown that in cap-to-cap arrangements the voltage interval between spikes is less than when the electrodes are in a translocation arrangement. Any spikes that occur when the voltage interval becomes large can be taken as not occurring during a period of oscillation in the sweep, instead they occur infrequently and randomly during the sweep.\par   

Reducing the frequency of the voltage sweep (Fig.~\ref{fig:slow}) also has the effect of removing the current oscillations.\par

\clearpage

\section{Mathematical Model of Mushroom Memfractance}
\label{sec:model}

Here we report the I-V characteristics of grey oyster fungi \emph{Pleurotus ostreatus} fruit bodies. It is evident from the results that grey oyster fungi display memristive behaviour.\par

Although the fruit bodies typically do not demonstrate the ``pinching" property of an ideal memristor~\cite{chua2014if}, it can be clearly seen that the biological matter exhibits memory properties when the electrical potential across the substrate is swept. A positive sweep yields a higher magnitude current when the applied voltage is positive; and a smaller magnitude current when the applied voltage is negative.\par 
\begin{figure}
    \centering
    \includegraphics[width=0.7\textwidth, trim=4 4 4 4,clip]{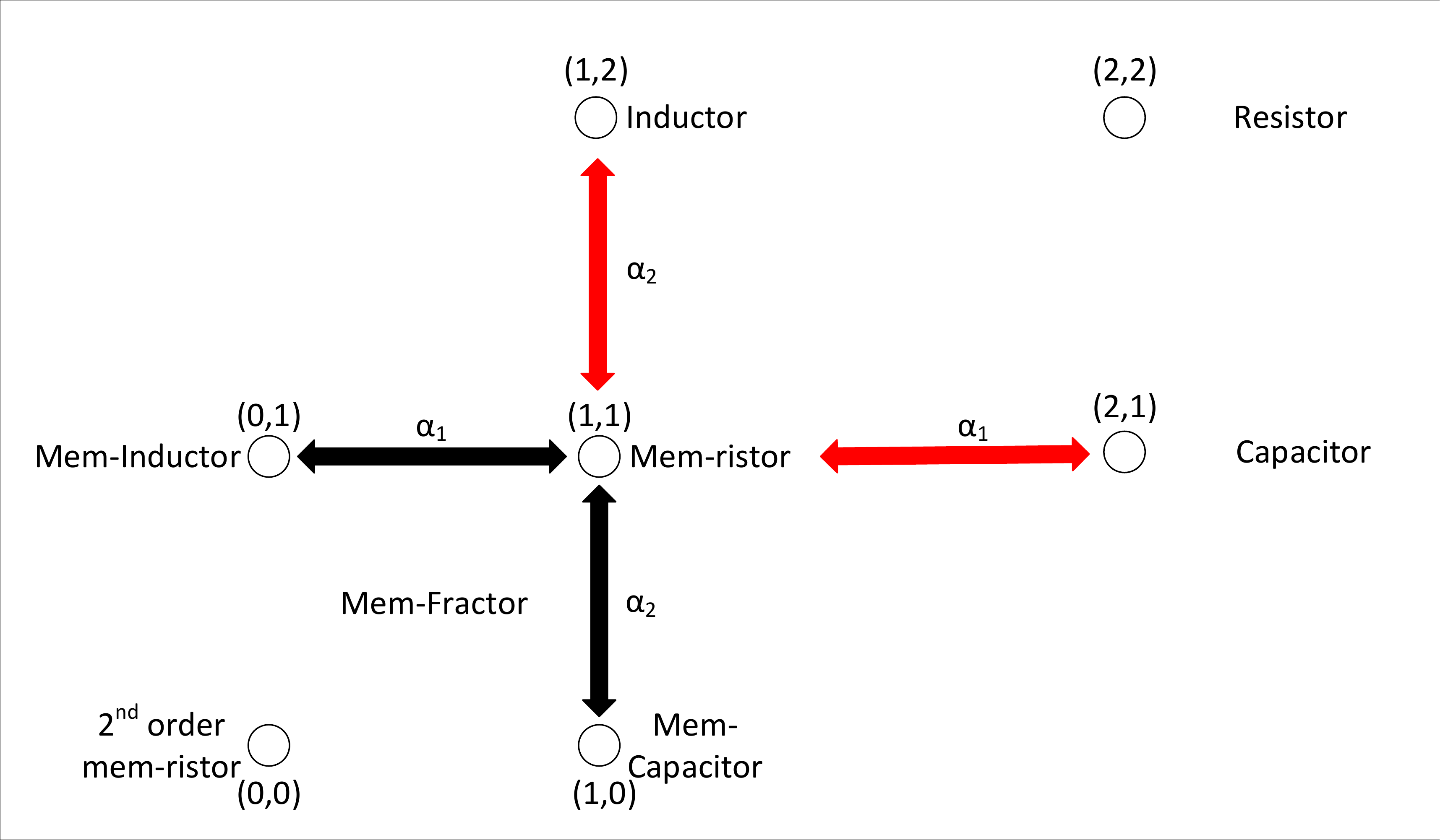}
    \caption{Non-binary solution space showing mem-fractive properties of a memory element}
    \label{fig:mem-fractor_space}
\end{figure}

\added[id=AB]{Fractional Order Memory Elements (FOME) are proposed as a combination of Fractional Order Mem-Capacitors (FOMC) and Fractional Order Mem-Inductors (FOMI)~\cite{Abdelouahab2014memfractance}. The FOME (\ref{eq:FOME_model}) is based on the generalised Ohm's law and parameterised as follows: $\alpha_1, \alpha_2$ are arbitrary real numbers --- it is proposed that  $0 \geq \alpha_1, \alpha_2 \leq 1$ models the solution space by~\cite{Khalil2019Fractional}, $F_M^{\alpha_1, \alpha_2}$ is the memfractance, $q(t)$ is the time dependent charge, $\varphi(t)$ is the time dependent flux. Therefore, the memfractance ($F_M^{\alpha_1,\alpha_2}$) is an interpolation between four points: $MC$ --- mem-capacitance, $R_M$ --- memristor, $MI$ --- mem-inductance, and $R_{2\,M}$ --- the second order memristor. Full derivations for the generalised FOME model are given by~\cite{Abdelouahab2014memfractance, Khalil2019Fractional}. The definition of memfractance can be straightforward generalised  to any value of $\alpha_1$, $\alpha_2$ (see [1, Fig. 27]).}

\begin{equation}
    D_t^{\alpha_1} \varphi (t) = F_M^{\alpha_1,\alpha_2}(t) D_t^{\alpha_2} q(t)
    \label{eq:FOME_model}
\end{equation}

\added[id=AB]{The appearance of characteristics from various memory elements in the fungal I-V curves supports the assertion that the fungal is a mem-fractor where $\alpha_1$ and $\alpha_2$ are both greater than 0 and less than 2.\par}

\begin{figure}[!tpb]
    \centering
    \includegraphics[width=0.8\textwidth]{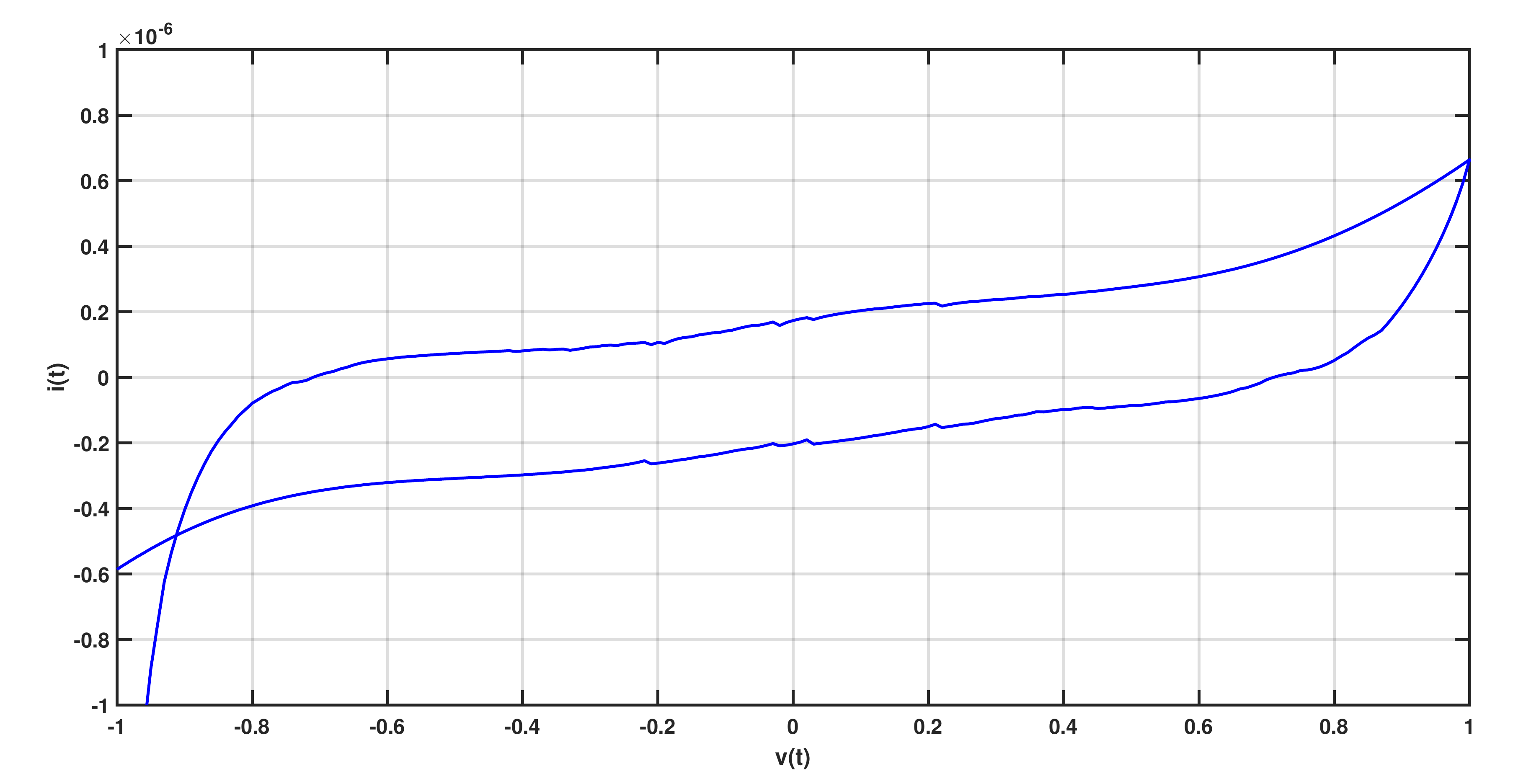}
    \caption{Raw data from average fruit bodies I-V cyclic voltammetry performed over -1 V to 1 V. Stem-to-cap electrode placement.}
    \label{fig:rms_1}
\end{figure}

\added[id=RL]{There is no biological reason for memfractance of Ooyster fungi fruit bodies with stem to cap electrodes, be a usual closed formula. Therefore, one can get only a mathematical approximation of this function. In this section, we propose two alternatives to obtain the best approximation for memfractance in the case of average fruit bodies I-V characteristics for cyclic voltammetry of Fig.~\ref{fig:2V_average_stem_to_cap} (Fig.~\ref{fig:rms_1}). }

\subsection{Approximation by polynomial on the whole interval of voltage}  

\begin{figure}[!tpb]
    \centering
    \includegraphics[width=0.8\textwidth]{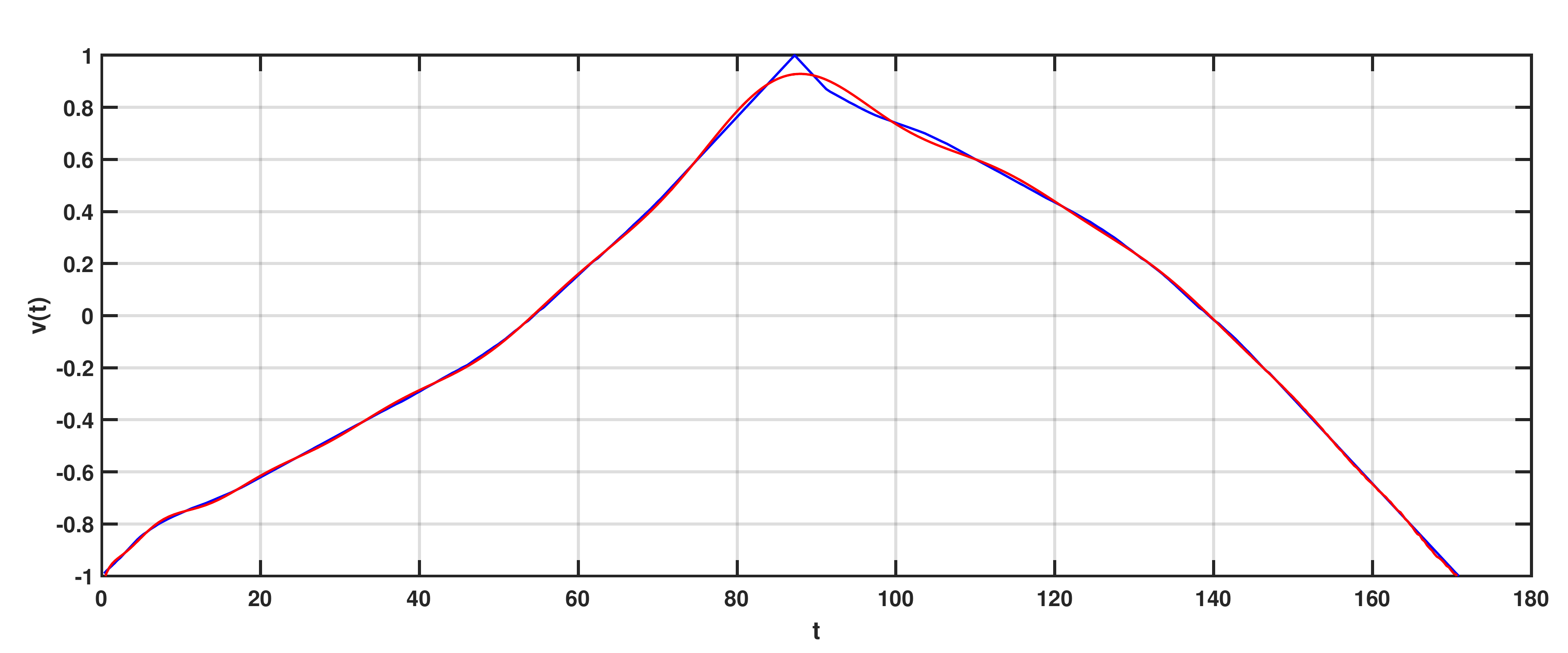}
    \caption{Voltage versus time and its approximation by a 24-degree polynomial}
    \label{fig:rms_2}
\end{figure}

\added[id=RL]{Raw data include the time, voltage and intensity of each reading. There are 171 readings for each run.
The process of these data, in order to obtain a mathematical approximation of memfractance, in the first alternative, takes 4 steps as follows.
First step: approximate $v(t)$ by a twenty-four-degree polynomial (Fig.~\ref{fig:rms_2}) whose coefficients are given in Tab,~\ref{tab:fifth_order_v}.}

\begin{equation}
    v(t) \approx P(t) = \sum_{j=0}^{j=24}a_jt^j
\end{equation}

\begin{table}[!tpb]
    \centering
        \caption{\textbf{Coefficient of P(t)}}
        \begin{tabular}{|c|c|c|c|}
            \hline
            $a_0$ & 2.36810109946699e-43 & $a_{13}$ & 1.18302125464207e-12 \\ \hline
            $a_1$ & -4.78342788514078e-40 & $a_{14}$ & -6.72265349925510e-11 \\ \hline
            $a_2$ & 4.45025298649318e-37 & $a_{15}$ &  2.90458838155410e-09 \\ \hline
            $a_3$ & -2.52211206380669e-34 & $a_{16}$ & -9.51752589043893e-08 \\ \hline
            $a_4$ & 9.68672841708898e-32 & $a_{17}$ & 2.33484036114612e-06 \\ \hline
            $a_5$ & -2.64464369703488e-29 & $a_{18}$ & -4.19159536470121e-05 \\ \hline
            $a_6$ & 5.19611819410190e-27 & $a_{19}$ & 0.000531866967507868 \\ \hline
            $a_7$ & -7.12198496974121e-25 & $a_{20}$ & -0.00453232038841485 \\ \hline
            $a_8$ & 5.80230108481181e-23 & $a_{21}$ & 0.0240895989682110 \\ \hline
            $a_9$ & 1.59013702626457e-22 & $a_{22}$ & -0.0726485498614107 \\ \hline
            $a_{10}$ & -8.60157726907686e-19 & $a_{23}$ & 0.135299293073760 \\ \hline
            $a_{11}$ & 1.48292987584698e-16 & $a_{24}$ & -1.04736115240006\\ \hline
            $a_{12}$ & -1.56317950862153e-14 & & \\ \hline
        \end{tabular}
    \label{tab:fifth_order_v}
\end{table}

\begin{table}[!tpb]
    \centering
    \caption{\textbf{Goodness of fit}}
    \begin{tabular}{|c|c|c|}
         \hline
        Sum of squared estimate of errors &  $SSE = \sum_{j=1}^{j=n}(v_j - \hat{v}_j)^2$ & 0.0680517563652170 \\ \hline
        Sum of squared residuals & $SSR = \sum_{j=1}^{j=n}(\hat{v}_j - \overline{v})^2$  & 133.688517134422\\  \hline
        Sum of square total & SST = SSE + SSR &  133.756568890787 \\ \hline
        Coefficient of determination  & $R-\text{square} = \frac{SSR}{SST}$ & 0.999491226809049\\ \hline
    \end{tabular}
    \label{tab:goodness}
\end{table}

\begin{figure}[!tpb]
    \centering
    \includegraphics[width=0.8\textwidth]{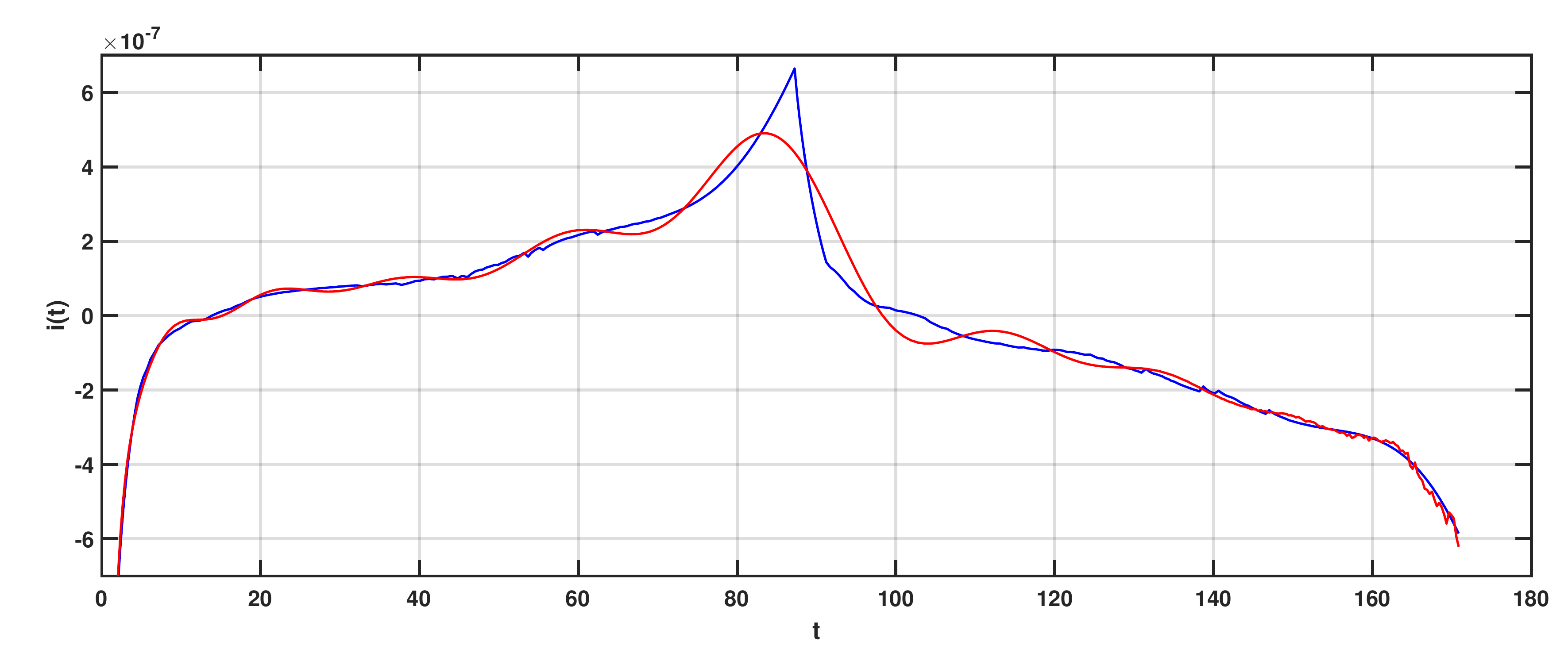}
    \caption{Current versus time and its approximation by a 24 degree polynomial }
    \label{fig:rms_3}
\end{figure}

\added[id=RL]{The polynomial fits very well the experimental voltage curve, as the statistical indexes show in Tab,~\ref{tab:goodness}.
Step 2: in the same way approximate the current $i(t)$ using a twenty-four-degree polynomial (Fig.~\ref{fig:rms_3}) whose coefficients are given in Table~\ref{tab:fifth_order_i}.}

\begin{equation}
    i(t) \approx Q(t) = \sum_{j=0}^{j=24}b_jt^j
\end{equation}

\begin{table}[!tpb]
    \centering
        \caption{\textbf{Coefficient of Q(t)}}
        \begin{tabular}{|c|c|c|c|}
            \hline
            $b_0$ & 8.73846352218898e-49 & $b_{13}$ & 4.25821973203331e-18 \\ \hline
            $b_1$ & -1.71535341852628e-45 & $b_{14}$ & -2.42471413376463e-16 \\ \hline
            $b_2$ & 1.55262364796050e-42 & $b_{15}$ & 1.06306849079070e-14 \\ \hline
            $b_3$ & -8.56384614589988e-40 & $b_{16}$ & -3.58289124918788e-13 \\ \hline
            $b_4$ & 3.19831491989559e-37 & $b_{17}$ & 9.19419585703268e-12 \\ \hline
            $b_5$ & -8.46294727047340e-35 & $b_{18}$ & -1.76692818009608e-10 \\ \hline
            $b_6$ & 1.59708666114599e-32 & $b_{19}$ & 2.48346182702953e-09 \\ \hline
            $b_7$ & -2.04593370725626e-30 & $b_{20}$ & -2.47326661661364e-08 \\ \hline
            $b_8$ & 1.36036443304302e-28 & $b_{21}$ & 1.67584032221916e-07 \\ \hline
            $b_9$ & 8.14484000064489e-27 & $b_{22}$ & -7.34738169887512e-07 \\ \hline
            $b_{10}$ & -3.66183256804588e-24 & $b_{23}$ & 1.95479195837707e-06 \\ \hline
            $b_{11}$ & 5.61870303550308e-22 & $b_{24}$ & -2.69478636561017e-06 \\ \hline
            $b_{12}$ & -5.69947824465678e-20 & & \\ \hline
        \end{tabular}
    \label{tab:fifth_order_i}
\end{table}

\begin{table}[!tpb]
    \centering
    \caption{\textbf{Goodness of fit}}
    \begin{tabular}{|c|c|}
        \hline
        Sum of squared estimate of errors & 5.84247524503151e-13 \\ \hline
        Sum of squared residuals &  4.07366051979587e-11 \\ \hline 
        Sum of square total & 4.13208527224619e-11 \\ \hline 
        Coefficient of determination & 0.985860709883522 \\ \hline 
    \end{tabular}
    \label{tab:goodness2}
\end{table}

\added[id=RL]{Again, the polynomial fits well the experimental intensity curve, as displayed in Table~\ref{tab:goodness2}.
Step 3: From (1) used under the following form $D_t^{\alpha_2}q(t) \neq 0$.}

\begin{equation}
    F_M^{\alpha_1,\alpha_2}(t) = \frac{D_t^{\alpha_1}\varphi{t}}{D_t^{\alpha_2}q(t)}
\end{equation}

\added[id=RL]{and the Rieman-Liouville fractional derivative defined by~\cite{FDEPodlubny1999}}

\begin{equation}
    {^{RL}_{0}}D_{t}^{\alpha}f(t) = \frac{1}{\Gamma(m-\alpha)}\frac{d^m}{dt^m}\int_{0}^{t}(t-s)^{m - \alpha - 1}f(s)ds\text{, m - 1} < \alpha < m  
\end{equation}

\added[id=RL]{together with the formula for the power function}

\begin{equation}
    {^{RL}_{0}}D_t^{\alpha}\left(a t^{\beta}\right) = \frac{a \Gamma(\beta + 1)}{\Gamma (\beta - \alpha + 1)}t^{\beta - \alpha}\text{, }\beta > -1, \alpha > 0,
\end{equation}

\added[id=RL]{we obtain the closed formula of $F_M^{\alpha_1,\alpha_2}(t)$, approximation of the true biological memfractance of the Oyster mushroom}

\begin{equation}
   F_M^{\alpha_1, \alpha_2}(t) = \frac{D_t^{\alpha_1}\varphi(t)}{D_t^{\alpha_2}\varphi(t)} = \frac{{^{RL}_{0}}D_t^{\alpha_1}\sum_{j=0}^{j=24}\frac{a_j}{j+1}t^{j+1}}{{^{RL}_{0}}D_{t}^{\alpha_2}\sum_{j=0}^{j=24}\frac{b_j}{j+1}t^{j+1}} = \frac{\sum_{j=0}^{j=24}\frac{a_j\Gamma(j+1)}{\Gamma(j+2-\alpha_1)}t^{j+1-\alpha_1}}{\sum_{j=0}^{j=24}\frac{b_j\Gamma(j+1)}{\Gamma(j+2-\alpha_2)}t^{j+1-\alpha_2}}
\end{equation}

\begin{figure}
    \centering
    \includegraphics[width=0.8\textwidth]{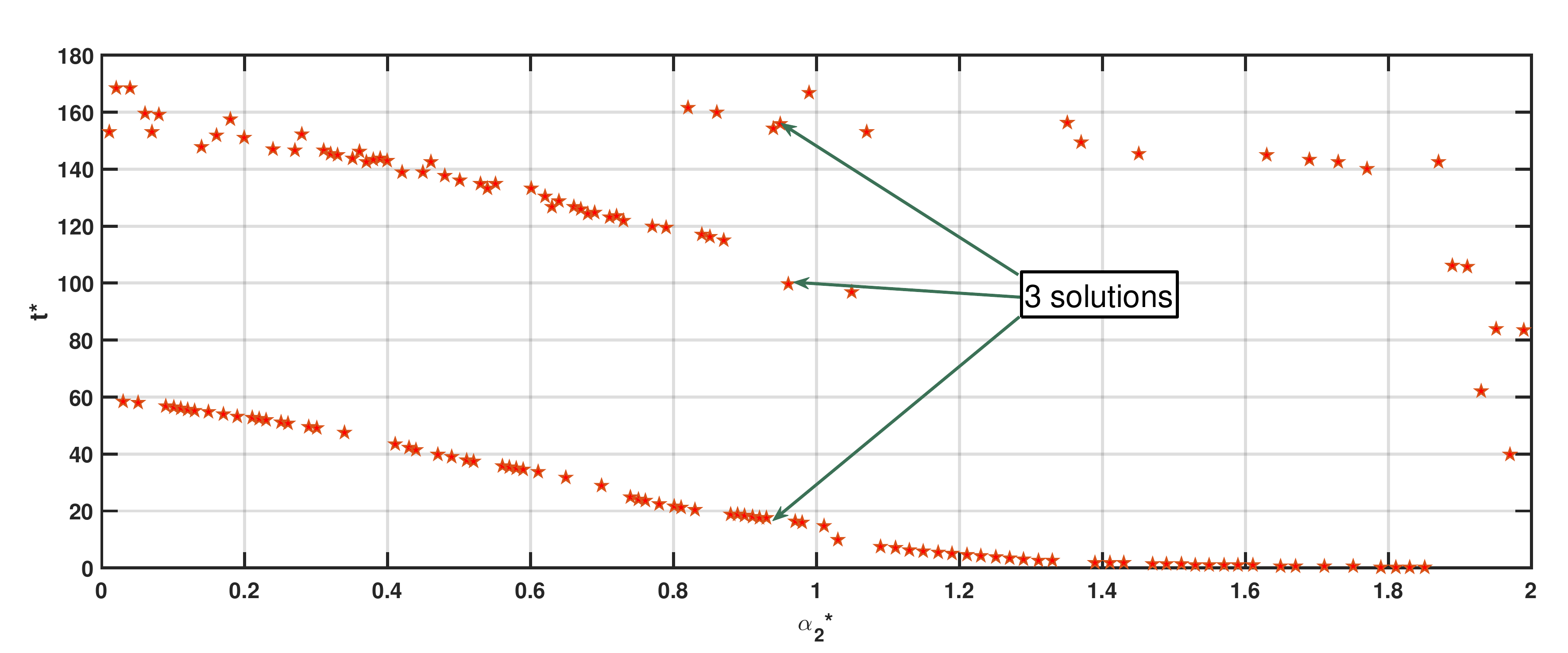}
    \caption{Zeros $t^*(\alpha_2)$ of the denominator of $F_M^{\alpha_1,\alpha_2}(t)$.}
    \label{fig:fig14}
\end{figure}

\added[id=RL]{Step 4 choice of parameter $\alpha_1$ and $\alpha_2$: We are looking for the best value of these parameters in the range $(\alpha_1,\alpha_2)\in[0,2]^2$. In this goal, we are considering first the singularities of $F_M^{\alpha_1,\alpha_2}(t)$ in order to avoid their existence, using suitable values of the parameters. Secondly, we will choose the most regular approximation. 
We compute numerically, the values $t^*(\alpha_2)$  which vanish the denominator of $F_M^{\alpha_1,\alpha_2}(t)$ (Fig.~\ref{fig:fig14}).}

\begin{figure}
    \centering
    \includegraphics[width=0.8\textwidth]{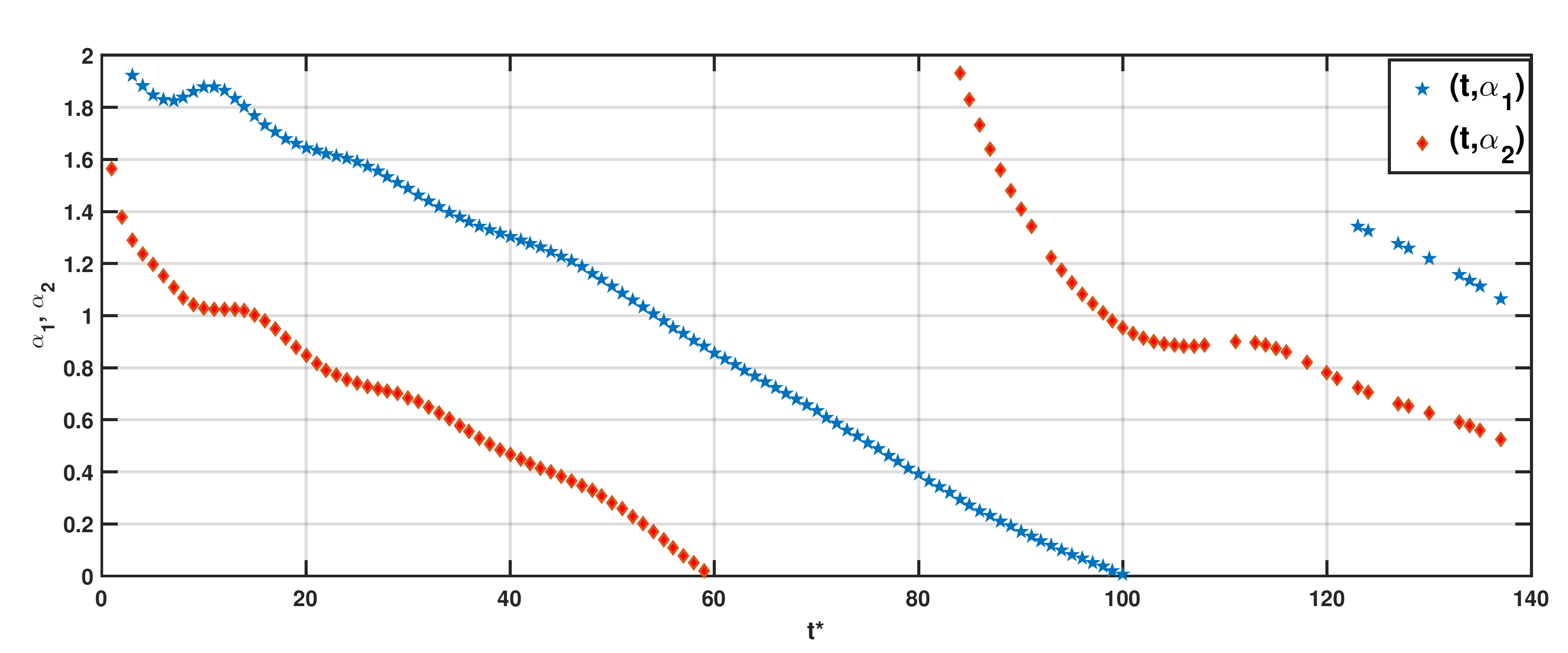}
    \caption{Zeros $t^*(\alpha_2)$ of $F_M^{\alpha_1,\alpha_2}(t)$ denominator (red dots), and zeros $t^*(\alpha_1)$ of the numerator (blue dots).}
    \label{fig:fig15}
\end{figure}

\begin{figure}
    \centering
    \includegraphics[width=0.8\textwidth]{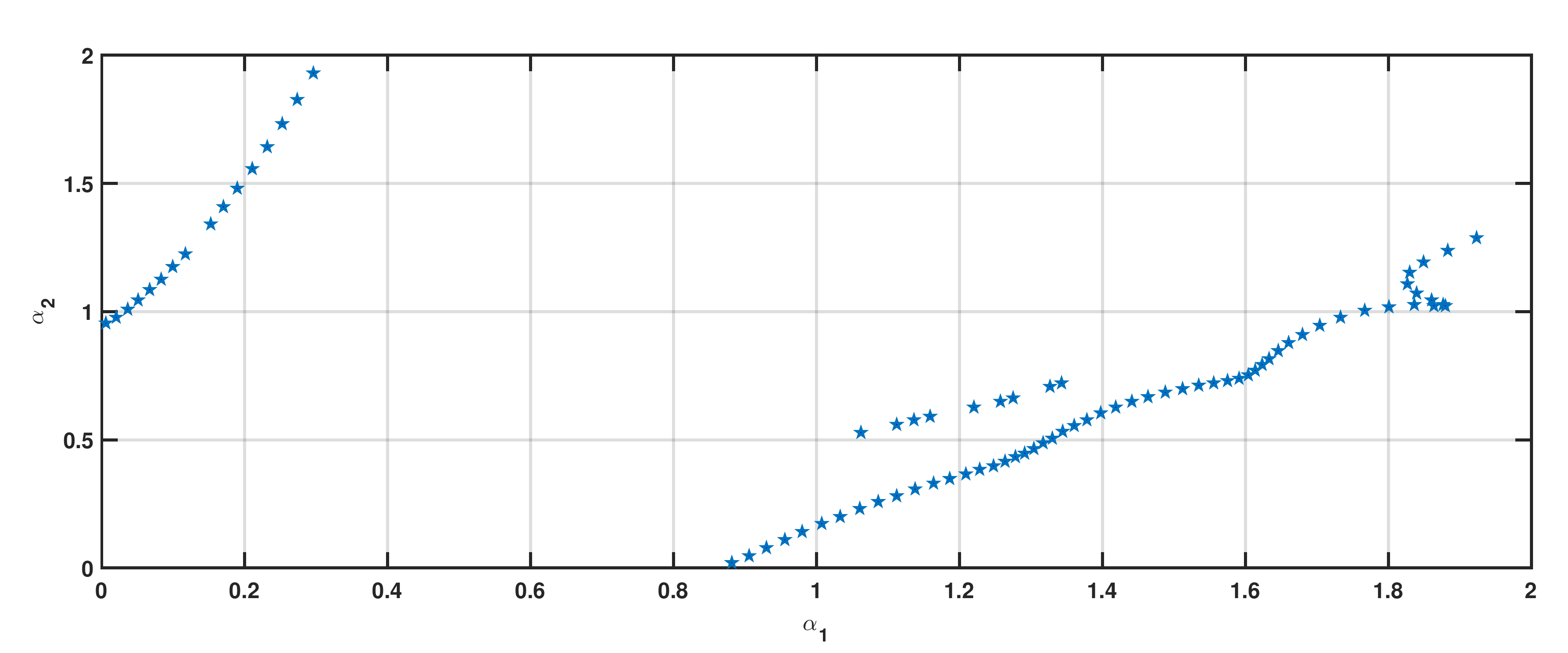}
    \caption{Values of $(\alpha_1,\alpha_2)\in[0,2]^2$ for which the zeros $t^*(\alpha_2)$ of denominator of $F_M^{\alpha_1,\alpha_2}(t)$ correspond to the zeros $t^*(\alpha_1)$ of denominator.}
    \label{fig:fig16}
\end{figure}

\added[id=RL]{We observe one, two or three coexisting solutions depending on the value of $\alpha_2$. Moreover, there is no value of $\alpha_2$ without zero of the denominator. Therefore, in order to eliminate the singularities, we need to determine the couples $(\alpha_1,\alpha_2)\in[0,2]^2$, vanishing simultaneously denominator and numerator of $F_M^{\alpha_1,\alpha_2}(t)$ (Figs.~\ref{fig:fig15},\ref{fig:fig16}).}

\begin{figure}
    \centering
    \includegraphics[width=0.8\textwidth]{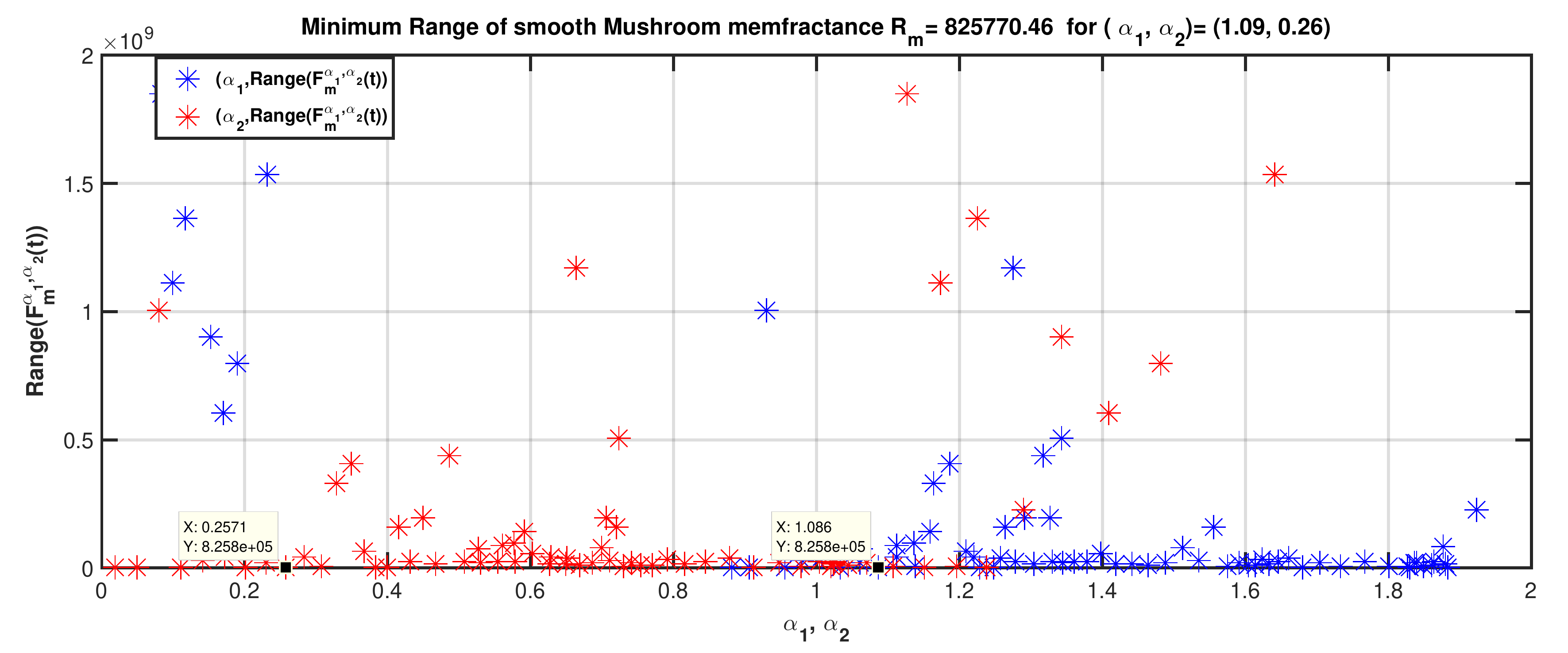}
    \caption{Values of range $(F_M^{\alpha_1,\alpha_2}(t))$  for  $(\alpha_1,\alpha_2)\in[0,2]^2$}
    \label{fig:fig17}
\end{figure}

\begin{figure}
    \centering
    \includegraphics[width=0.8\textwidth]{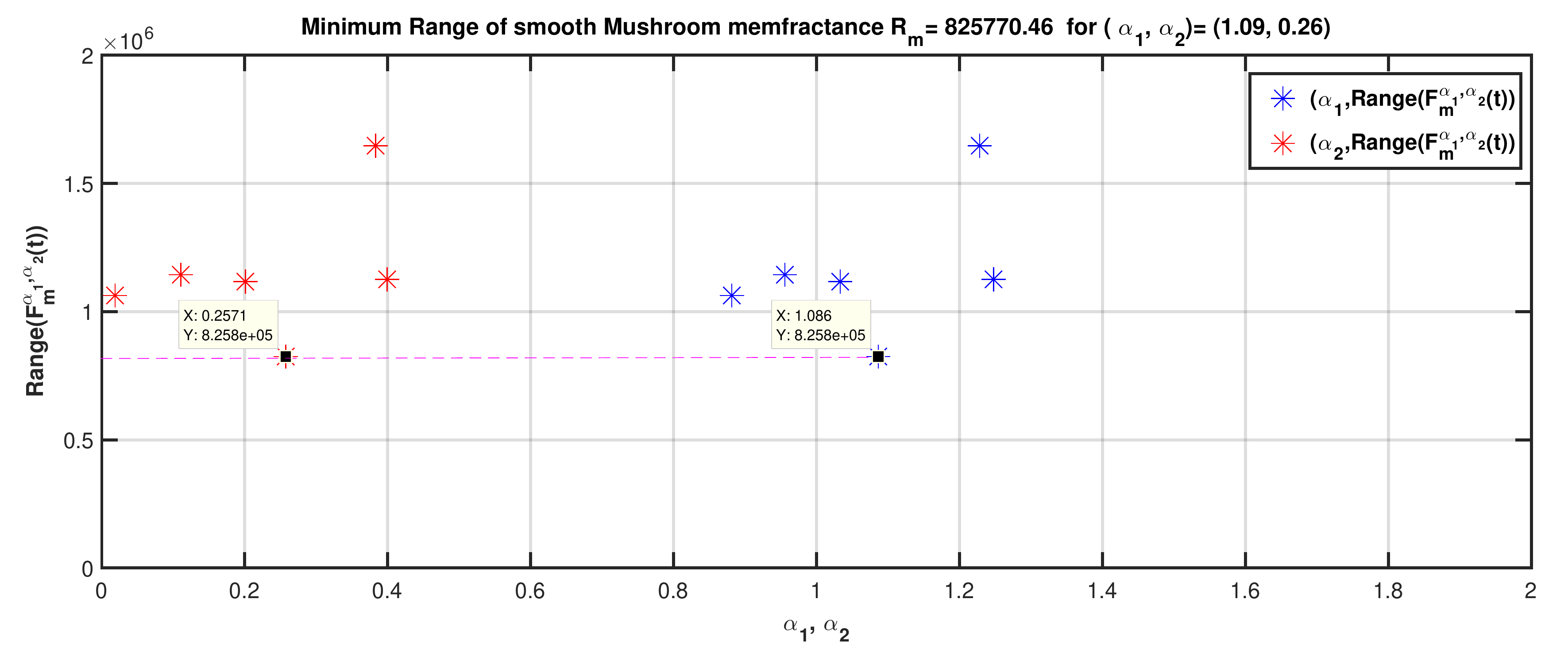}
    \caption{Magnification of Fig.~\ref{fig:fig17}.}
    \label{fig:fig18}
\end{figure}

\added[id=RL]{In the second part of step 4, we choose the most regular approximation. We consider that the most regular approximation is the one for which the function range $(F_M^{\alpha_1,\alpha_2}(t))$ is minimal (Figs.~\ref{fig:fig17},\ref{fig:fig18})}

\begin{equation}
    \text{range}\left(F_M^{\alpha_1,\alpha_2}(t)\right) = \max_{t \in [0,171]}\left(F_M^{\alpha_1,\alpha_2}(t)\right) - \min_{t \in [0,171]}\left(F_M^{\alpha_1,\alpha_2}(t)\right)
\end{equation}

\begin{figure}
    \centering
    \includegraphics[width=0.8\textwidth]{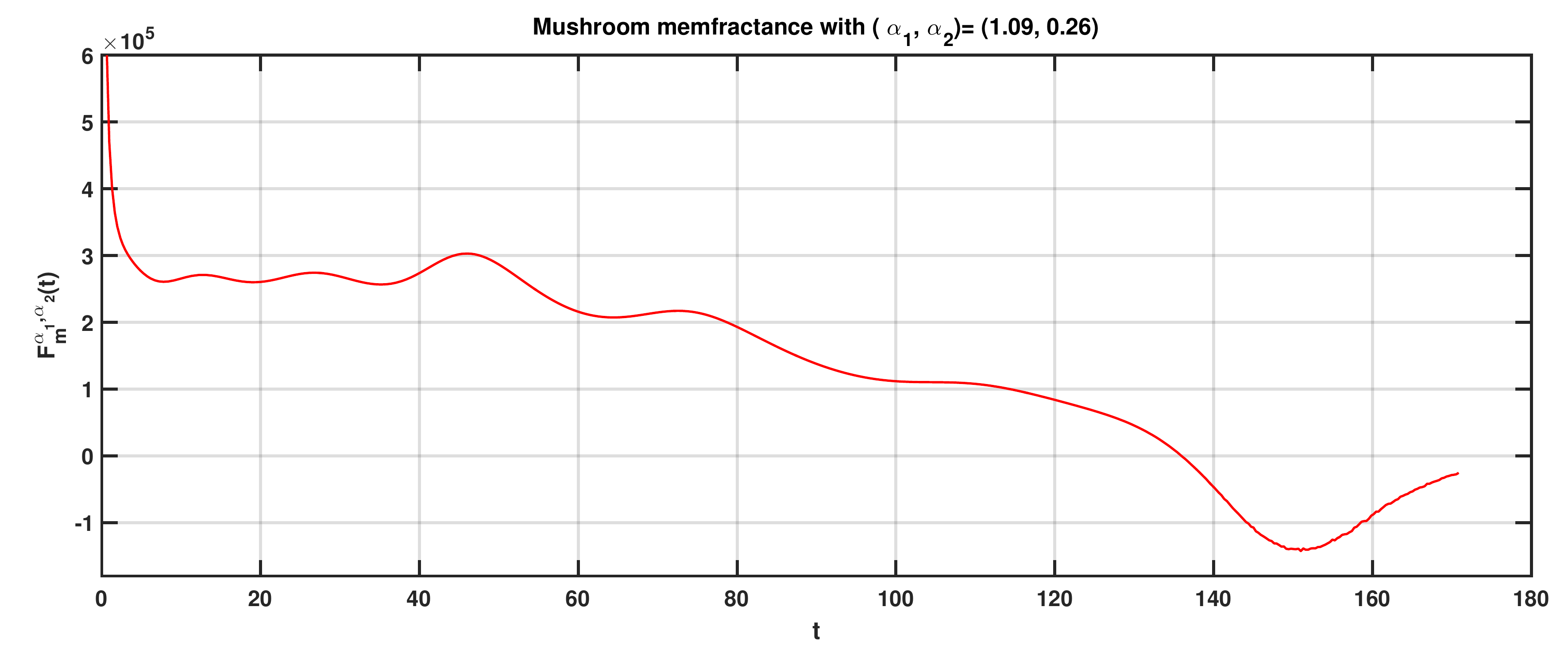}
    \caption{Memfractance for $(\alpha_1,\alpha_2)$ given in Table~\ref{tab:tab5}.}
    \label{fig:fig19}
\end{figure}

\added[id=RL]{From the numerical results, the best couple $(\alpha_1,\alpha_2)$ and the minimum range of $F_M^{\alpha_1,\alpha_2}(t)$ are given in Table~\ref{tab:tab5}, and the corresponding Memfractance is displayed in Fig.~\ref{fig:fig19}.}

\begin{table}[!tpb]
    \centering
    \caption{\textbf{Minimum $F_M^{\alpha_1,\alpha_2}(t)$}}
    \begin{tabular}{|c|c|c|}
        \hline
        \textbf{$\alpha_1$} & \textbf{$\alpha_2$} & \textbf{Minimum range $F_M^{\alpha_1,\alpha_2}(t)$} \\ \hline
        1.08642731 & 0.25709492 & 825770.46017259 \\ \hline
    \end{tabular}
    \label{tab:tab5}
\end{table}

\added[id=RL]{The value of $(\alpha_1,\alpha_2)$ given in Table~\ref{tab:tab5} belongs to the triangle of Fig.~\ref{fig:mem-fractor_space}, whose vertices are Memristor, Memcapacitor and Capacitor. Which means that Oyster mushroom fruit bodies with stem to cap electrodes, is like a mix of such basic electronic devices.}

\begin{figure}[!tpb]
    \centering
    \includegraphics[width=0.8\textwidth]{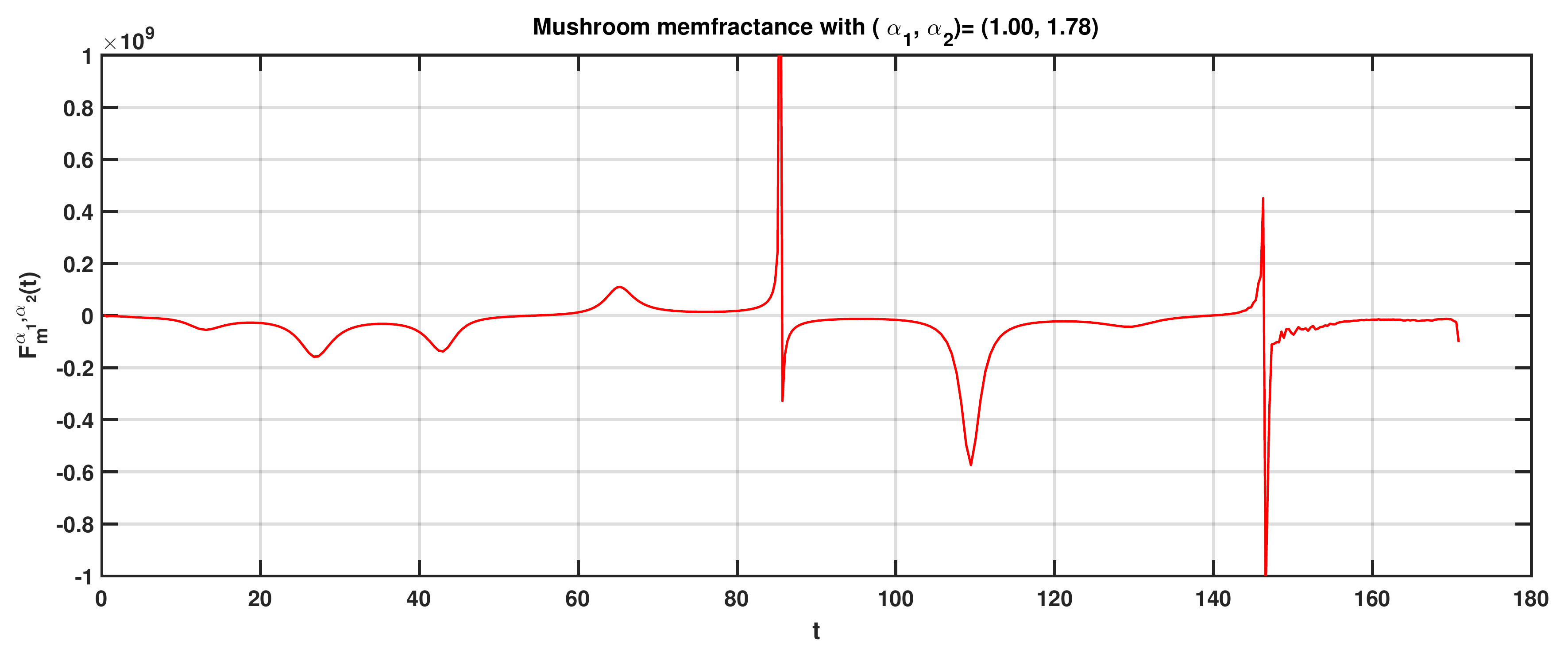}
    \caption{Memfractance with two singularities for $(\alpha_1,\alpha_2)=(1, 1.78348389322388)$.}
    \label{fig:fig20}
\end{figure}

\added[id=RL]{As a counter-example of our method for choosing the best possible Memfractance, Fig.~\ref{fig:fig20} displays, the Memfractance for a non-optimal couple $(\alpha_1,\alpha_2)=(1, 1.78348389322388)$ which presents two singularities.}

\subsection{Approximate cycling voltammetry}

\added[id=RL]{From the closed formula of $F_M^{\alpha_1^*,\alpha^*_2}(t)$ it is possible to retrieve the formula of the current function $i(t)$ using (1).}

\begin{equation}
    \begin{split}
    i(t)& = D_t^{1-\alpha_2}\left[\frac{D_t^{\alpha_1}\varphi(t)}{F_M^{\alpha_1,\alpha_2}(t)}\right] = D_t^{1-\alpha_2}\left[\frac{\sum_{j=0}^{j=24}\frac{a_j\Gamma(j+1)}{\Gamma(j+2-\alpha_1)}t^{j+1-\alpha_1}}{\frac{\sum_{j=0}^{j=24}\frac{a_j\Gamma(j+1)}{\Gamma(j+2-\alpha_1)}t^{j+1-\alpha_1}}{\sum_{j=0}^{j=24}\frac{b_j\Gamma(j+1)}{\Gamma(j+2-\alpha_2)}t^{j+1-\alpha_2}}}\right] \\
    & = D_t^{1-\alpha_2}\left[\sum_{j=0}^{j=24}\frac{b_j\Gamma(j+1)}{\Gamma(j+2-\alpha_2)}t^{j+1-\alpha_2}\right] \\
    & = \sum_{j=0}^{j=24}\frac{\Gamma(j+2-\alpha_2)b_j\Gamma(j+1)}{\Gamma(j+2-\alpha_2)\Gamma(j+1)}t^{j+1-\alpha_2-(1-\alpha_2)} \\
    & = \sum_{j=0}^{j=24}b_jt^j
    \end{split}
\end{equation}

\begin{figure}[!tpb]
    \centering
    \includegraphics[width=0.8\textwidth]{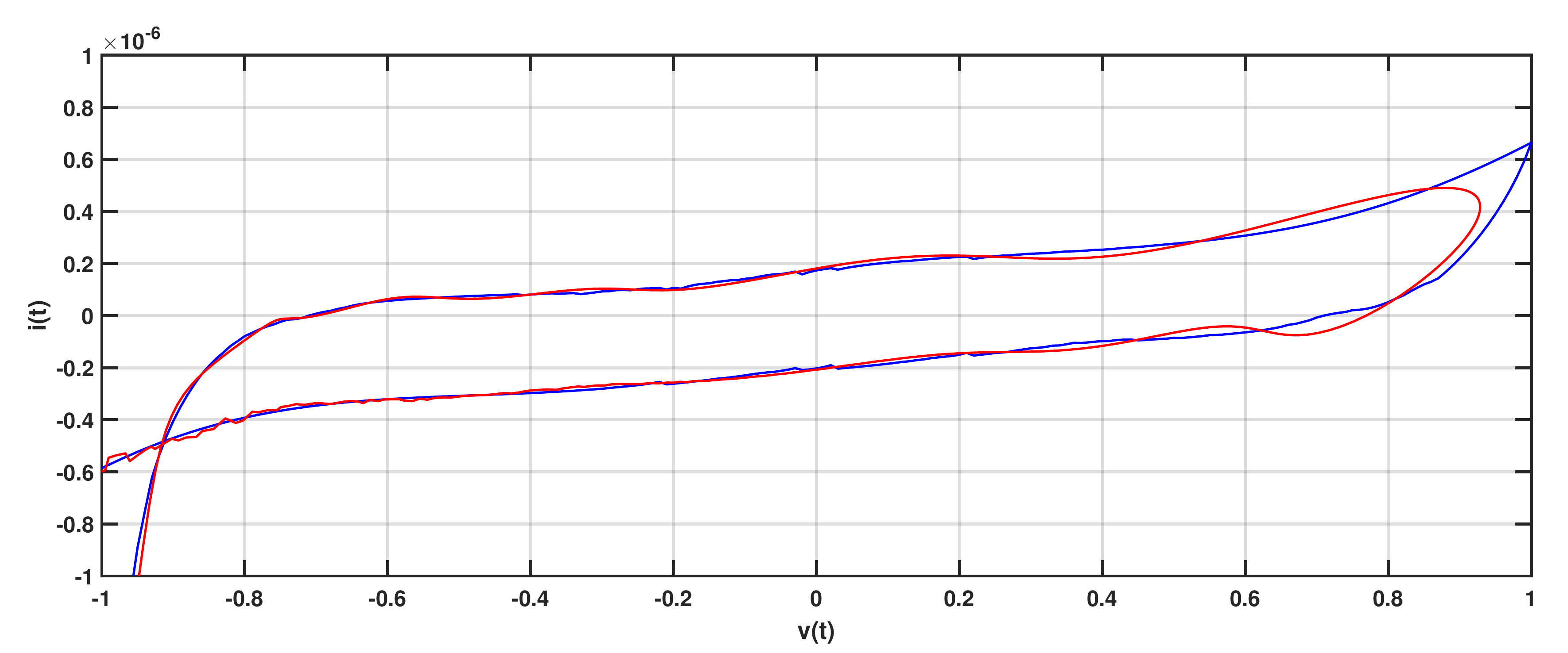}
    \caption{Comparison between average experimental data of cyclic voltammetry performed over -1\,V to 1\,V, Stem-to-cap electrode placement, and approximate values of $v(t)$ and $i(t)$.}
    \label{fig:fig21}
\end{figure}

\added[id=RL]{The comparison of average experimental data of cyclic voltammetry performed over -1\,V to 1\,V, Stem-to-cap electrode placement, and closed approximative formula is displayed in Fig.~\ref{fig:fig21}, showing a good agreement between both curves except near the maximum value of $v(t)$ and $i(t)$. }

\begin{figure}[!tpb]
    \centering
    \includegraphics[width=0.8\textwidth]{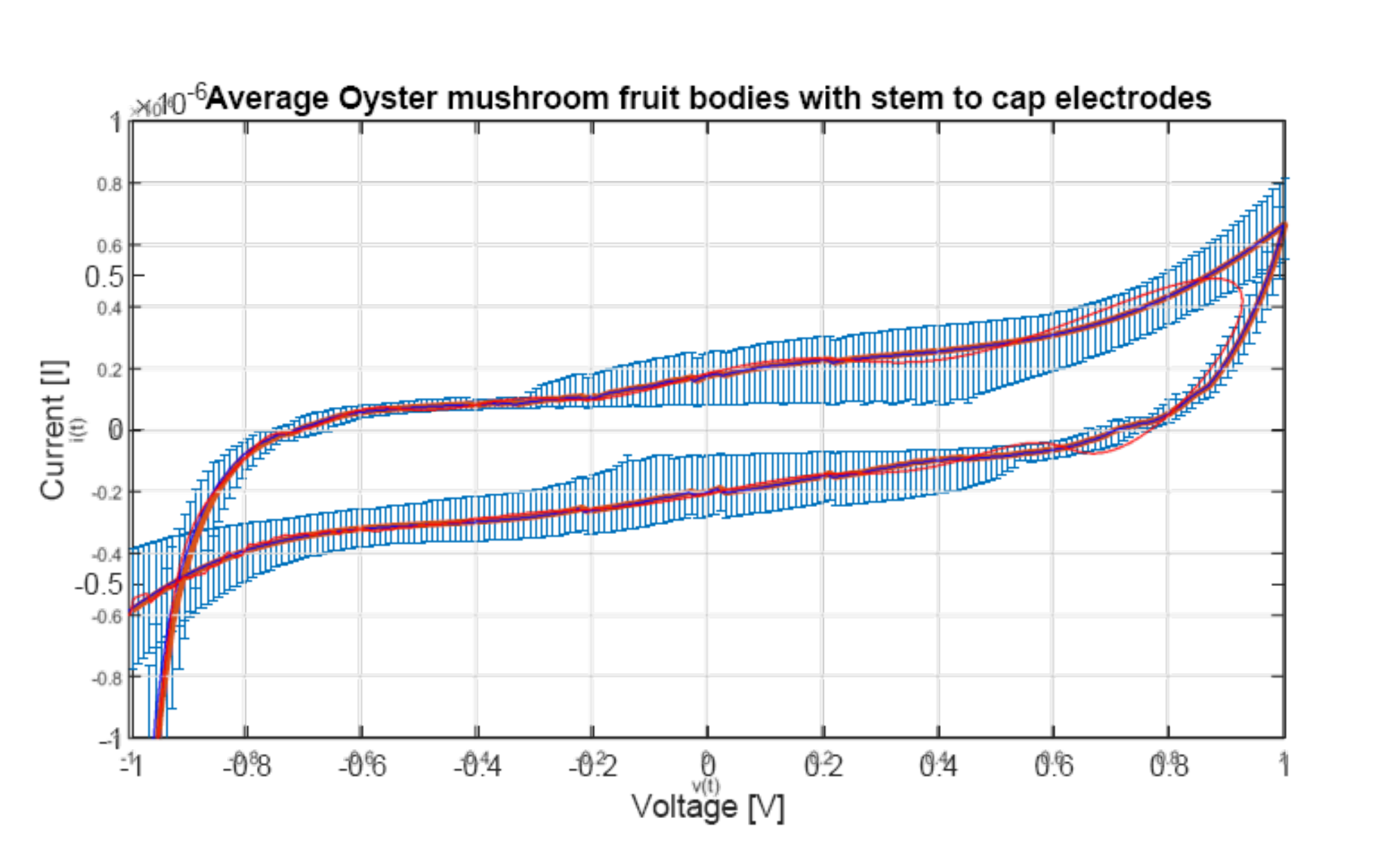}
    \caption{Both average experimental data curve and the curve computed from closed approximative formula are nested into the histogram of data of all runs.}
    \label{fig:fig22}
\end{figure}

\added[id=RL]{Figure~\ref{fig:fig22} shows that the curve computed from closed approximative formula belongs to the histogram of data of all runs.}

\added[id=RL]{The discrepancy between both curves is due to the method of approximation chosen in (2) and (3).}

\added[id=RL]{It is possible, as we show in the next subsection to improve the fitting of the approximated curve near the right hand-side vertex, using piecewise polynomial approximation of both $v(t)$ and $i(t)$.}

\subsection{Alternative approximation of the cycling voltammetry}

\added[id=RL]{Due to the way of conducting the experiments, the voltage curve presents a vertex, that means that the function $v(t)$ is non-differentiable for $T = 87.23747459$. In fact, the value of $T$ is the average value of the non-differentiable points for the 20 runs.}

\added[id=RL]{In this alternative approximation, we follow the same 4 steps as in 4.1, changing the approximation by a twenty-four-degree polynomial to an approximation by a 2-piecewise fifth-degree-polynomial, for both $v(t)$ and $i(t)$.}

\begin{figure}[!tpb]
    \centering
    \includegraphics[width=0.8\textwidth]{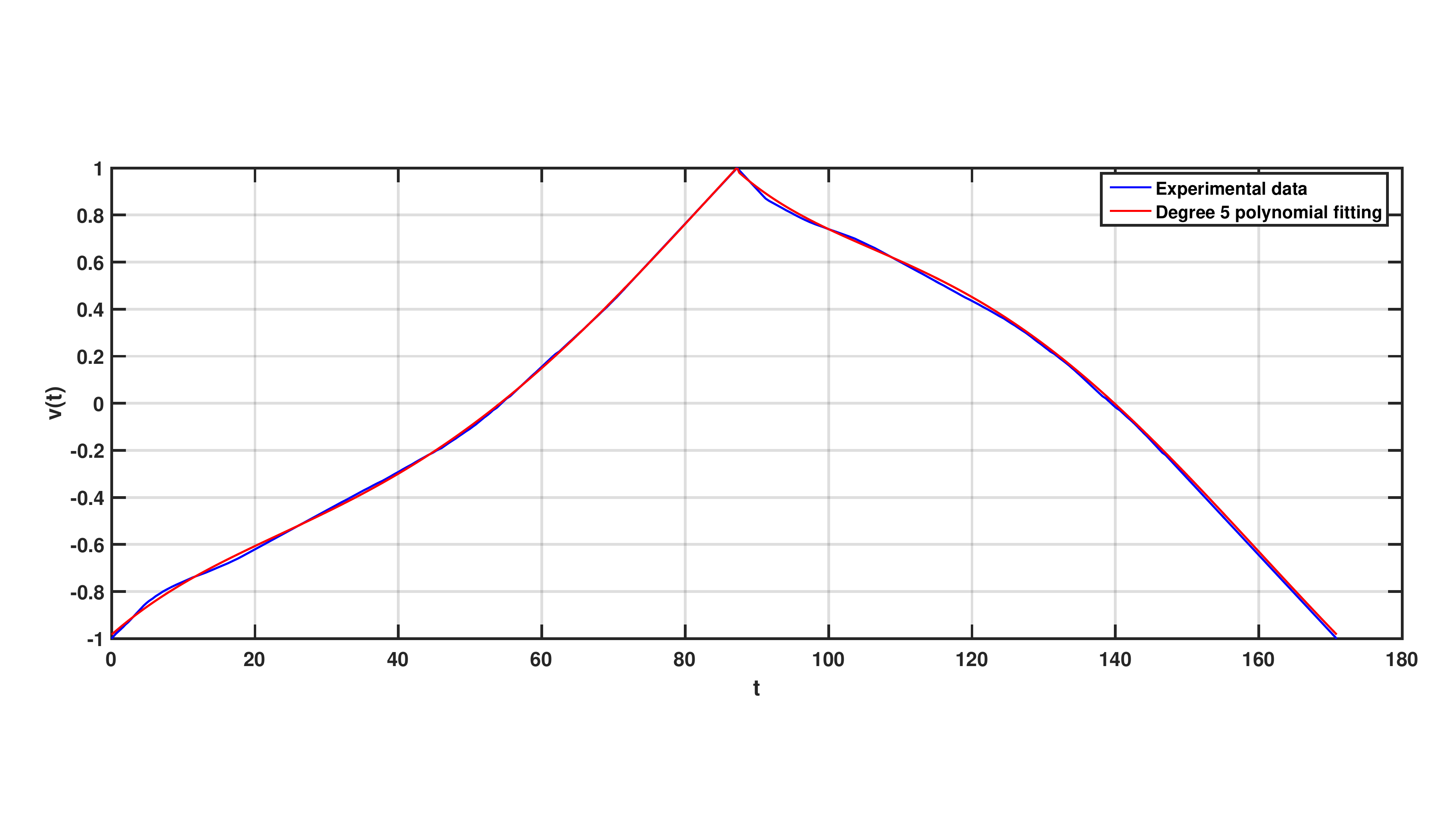}
    \caption{Voltage versus time and its approximation by 2-piecewise fifth degree polynomial }
    \label{fig:fig23}
\end{figure}

\added[id=RL]{First step: approximation of $v(t)$ by a 2-piecewise fifth-degree-polynomial (Fig.~\ref{fig:fig23}) whose coefficients are given in Table~\ref{tab:tab6}.}

\begin{equation}
    v(t) = 
            \begin{cases}
            P_1(t) = a_0 + a_1t + a_2t^2 + a_3t^3 + a_4t^4 + a_5t^5\text{, for } 0 \leq t \leq T \\
            P_2(t) = a^{\prime}_0 + a^{\prime}_1t + a^{\prime}_2t^2 + a^{\prime}_3t^3 + a^{\prime}_4t^4 + a^{\prime}_5t^5\text{, for T} \leq t < 171
            \end{cases}
\end{equation}

\begin{table}[!tpb]
    \centering
    \caption{\textbf{Coefficient for i(t)}}
    \begin{tabular}{|c|c|c|c|}
    \hline
        Coefficient & Value for $0\leq t \leq T$ & Coefficient & Value for $T \leq t < T $ \\ \hline
        $a_0$ & -0.98299 & $a^{\prime}_0$ & 37.16955 \\ \hline 
        $a_1$ & 0.02665 & $a^{\prime}_1$ & -1.2986 \\ \hline
        $a_2$ & -5.91565 E -4 & $a^{\prime}_2$ & 0.01826  \\ \hline
        $a_3$ & 1.12211 E -5 & $a^{\prime}_3$ & -1.25146 E -4 \\ \hline
        $a_4$ & -6.28483 E -8 & $a^{\prime}_4$ & 4.12302 E -7 \\ \hline
        $a_5$ & 6.9675 E -11 & $a^{\prime}_5$ & -5.25359 E-19 \\ \hline
    \end{tabular}
    \label{tab:tab6}
\end{table}

\added[id=RL]{The flux is obtained integrating $v(t)$ versus time.}

\begin{equation}
    \varphi(t) = \begin{cases}
                IP_1(t) = a_0t + \frac{a_1}{2}t^2 + \frac{a_2}{3}t^3 + \frac{a_3}{4}t^4 + \frac{a_4}{5}t^5 + \frac{a_5}{6}t^6\text{, for }0 \leq t \leq T \\
                IP_2(t) = a^{\prime}_0t + \frac{a^{\prime}_1}{2}t^2 + \frac{a^{\prime}_2}{3}t^3 + \frac{a^{\prime}_3}{4}t^4 + \frac{a^{\prime}_4}{5}t^5 + \frac{a^{\prime}_5}{6}t^6\text{, for }T \leq t < 171 \\
                \end{cases}
\end{equation}

\added[id=RL]{The polynomial fits very well the experimental voltage curve, as the statistical indexes show in Table~\ref{tab:tab7}.}

\begin{table}[!tpb]
    \centering
    \caption{\textbf{Goodness of fit}}
    \begin{tabular}{|c|c|c|}
    \hline
        Approximation &  $t < T$ & $ t > T $ \\ \hline
        Coefficient of determination & 0.99983 & 0.9999 \\ \hline
    \end{tabular}
    \label{tab:tab7}
\end{table}

\begin{figure}
    \centering
    \includegraphics[width=0.8\textwidth]{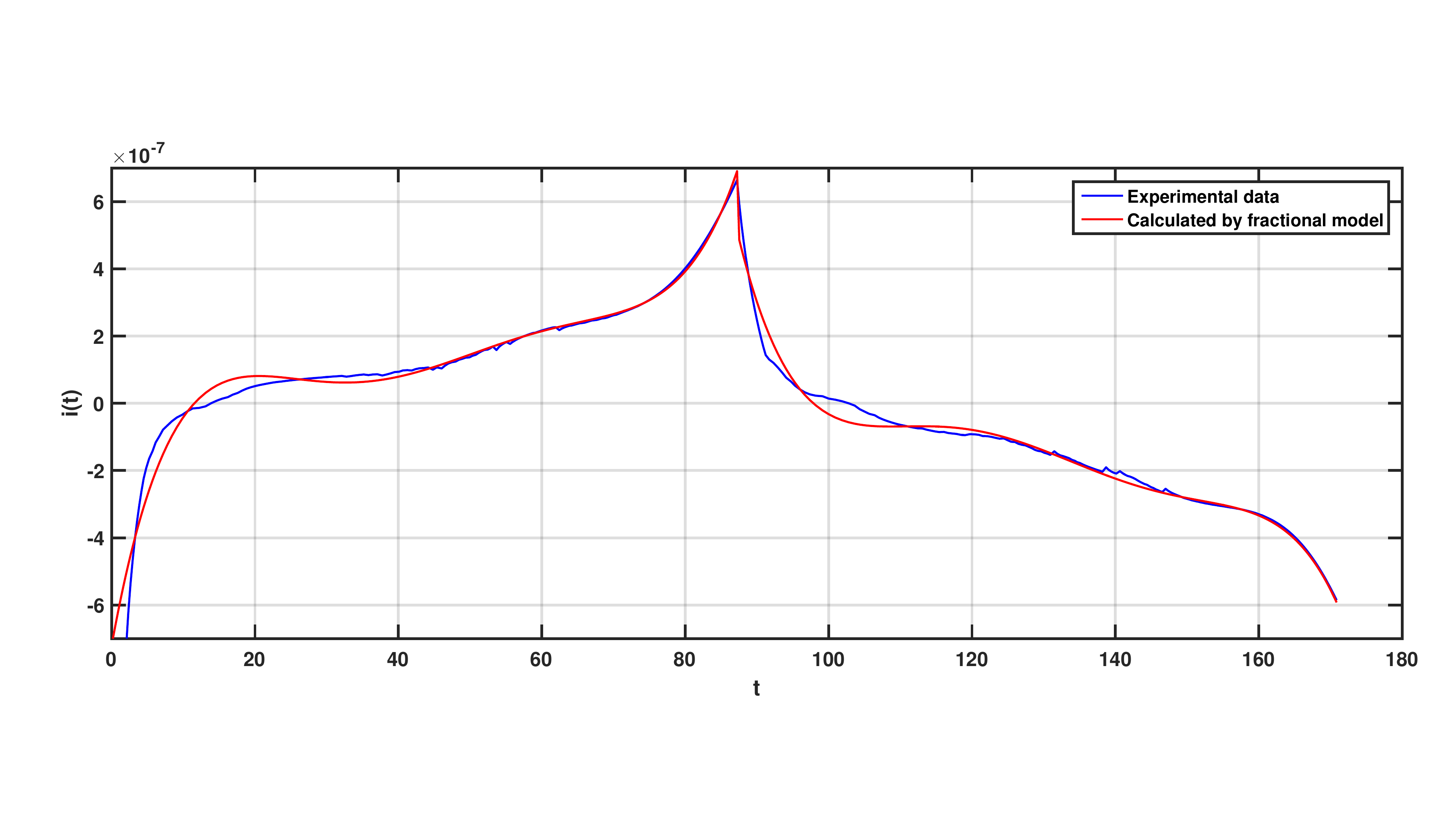}
    \caption{Current versus time and its approximation by 2-piecewise fifth degree polynomial }
    \label{fig:fig24}
\end{figure}

\added[id=RL]{Step 2: in the same way, one approximates the current $i(t)$ using a 2-piecewise fifth degree polynomial (Fig.~\ref{fig:fig24}) whose coefficients are given in Table~\ref{tab:tab8}.}

\begin{equation}
    i(t) = \begin{cases}
            P_3(t) = b_0 + b_1t + b_2t^2 + b_3t^3 + b_4t^4 + b_5t^5\text{, for }0 \leq t \leq T \\
            P_4(t) = b^{\prime}_0 + b^{\prime}_1t + b^{\prime}_2t^2 + b^{\prime}_3t^3 + b^{\prime}_4t^4 + b^{\prime}_5t^5\text{, for }T \leq t < 171
            \end{cases}
\end{equation}

\begin{table}[!tpb]
    \centering
    \caption{\textbf{Coefficient for i(t)}}
    \begin{tabular}{|c|c|c|c|}
    \hline 
        Coefficient & Value for $0 \leq t \leq T$ & Coefficient & Value for $T \leq t < 171$ \\ \hline
        $b_0$ & -7.21418 E -7 & $b^{\prime}_0$ & 2.69466 E -4 \\ \hline
        $b_1$ & 1.11765 E -7 & $b^{\prime}_1$ & -1.05461 E -5 \\ \hline
        $b_2$ & -6.3792 E -9 & $b^{\prime}_2$ & 1.63678 E -7 \\ \hline
        $b_3$ & 1.57327 E -10 & $b^{\prime}_3$ & -1.25915 E -9 \\ \hline
        $b_4$ & -1.7745 E -12 & $b^{\prime}_4$ & 4.80107 E -12 \\ \hline
        $b_5$ & 7.52304 E -15 & $b^{\prime}_5$ & -7.26253 E-15\\ \hline
    \end{tabular}
    \label{tab:tab8}
\end{table}

\added[id=RL]{Again, the polynomial fits very well the experimental voltage curve, as the statistical indexes show in Table~\ref{tab:tab9}.}

\begin{table}[!tpb]
    \centering
    \caption{\textbf{Goodness of fit}}
    \begin{tabular}{|c|c|c|}
    \hline 
        Approximation &  $t < T$ & $t > T$\\ \hline 
        Coefficient of determination & 0.99171 & 0.98613 \\ \hline 
    \end{tabular}
    \label{tab:tab9}
\end{table}

\added[id=RL]{Therefore, the charge is given by}

\begin{equation}
    q(t) = \begin{cases} 
            IP_3(t) = b_0t + \frac{b_1}{2}t^2 + \frac{b_2}{3}t^3 + \frac{b_3}{4}t^4 + \frac{b_4}{5}t^5+ \frac{b_5}{6}t^6\text{, for } 0 \leq t \leq T \\
            IP_4(t) = b^{\prime}_0t + \frac{b^{\prime}_1}{2}t^2 + \frac{b^{\prime}_2}{3}t^3 + \frac{b^{\prime}_3}{4}t^4 + \frac{b^{\prime}_4}{5}t^5+ \frac{b^{\prime}_5}{6}t^6\text{, for } T \leq t < 171
            \end{cases}
\end{equation}

\added[id=RL]{Step 3: Following the same calculus as before with (4), one obtains}

\begin{equation}
    \text{for } 0 \leq t \leq T\text{, }F_M^{\alpha_1,\alpha_2}(t) = \frac{{^{RL}_{0}}D_t^{\alpha_1}\varphi(t)}{{^{RL}_{0}}D_t^{\alpha_2}q(t)} = \frac{{^{RL}_{0}}D_t^{\alpha_1}[IP_1(t)]}{{^{RL}_{0}D_t^{\alpha_2}[IP_3(t)]}} = \frac{\sum_{j = 1}^{j=5}\frac{a_j\Gamma(j+1)}{\Gamma(j+2-\alpha_1)}t^{j+1-\alpha_1}}{\sum_{j=0}^{j=5}\frac{b_j\Gamma(j+1)}{\Gamma(j+2-\alpha_2)}t^{j+1-\alpha_2}}
\end{equation}

\added[id=RL]{However, because fractional derivative has memory effect, for $T<t<171$, the formula is slightly more complicated}

\begin{equation}
    \begin{split}
    F_M^{\alpha_1,\alpha_2}(t) & = \frac{{^{RL}_{0}}D_t^{\alpha_1}\varphi(t)}{{^{RL}_{0}}D_t^{\alpha_2}q(t)} = \frac{\frac{1}{\Gamma(m_1-\alpha_1)}\frac{d^{m_1}}{dt^{m_1}}\int_{0}^{t}(t-s)^{m_1 - \alpha_1 - 1}\varphi(s)ds}{\frac{1}{\Gamma(m_2-\alpha_2)}\frac{d^{m_2}}{dt^{m_2}}\int_{0}^{t}(t-s)^{m_2-\alpha_2-1}q(s)ds}\text{, }m_1 - 1 < \alpha_1 < m_1\text{ and }m_2 - 1 < \alpha_2 < m_2 \\
    & = \frac{\frac{1}{\Gamma(m_1 - \alpha_1)}\frac{d^{m_1}}{dt^{m_1}}\left[ \int_{0}^{T}(t-s)^{m_1 - \alpha_1 - 1} IP_1(s)ds + \int_T^t(t-s)^{m_1 - \alpha_1 - 1} IP_2(s)ds \right]}{\frac{1}{\Gamma(m_2-\alpha_2)}\frac{d^{m_2}}{dt^{m_2}}\left[ \int_0^T(t-s)^{m_2 - \alpha_2 - 1}IP_3(s)ds + \int_{T}^{t}(t-s)^{m_2 - \alpha_2 - 1}IP_4(s)ds \right]}\\
    & = \frac{\frac{1}{\Gamma(m_1 - \alpha_1)}\frac{d^{m_1}}{dt^{m_1}} \sum_{j=0}^{j=5} \left[ \frac{a_j}{j+1}\int_{0}^{T}(t-s)^{m_1 - \alpha_1 - 1}s^{j+1}ds + \frac{a^{\prime}_j}{j+1}\int_{T}^{t}(t-s)^{m_1 - \alpha_1 - 1}s^{j+1}ds \right]}{ \frac{1}{\Gamma(m_2 - \alpha_2)} \frac{d^{m_2}}{dt^{m_2}} \sum_{j=0}{j=5}\left[ \frac{b_j}{j+1} \int_{0}^{T}(t-s)^{m_2 - \alpha_2 - 1}s^{j+1}ds + \frac{b^{\prime}_j}{j+1} \int_{T}^{t}(t-s)^{m_2 - \alpha_2 - 1} s^{j+1}ds \right] }
    \end{split}
\end{equation}

\added[id=RL]{Using integration by part repeatedly six times we obtain}
\begin{landscape}
\begin{equation}
    \begin{split}
        & F_M^{\alpha_1,\alpha_2}(t) \\
        & = \frac{ \frac{1}{\Gamma(m_1 - \alpha_1)} \frac{d^{m_1}}{dt^{m_1}} \sum_{j=0}^{j=5}\left[ \frac{a_j}{j+1} \left[ \sum_{k=0}^{k=j+1} \left[ \frac{-(j+1)!\Gamma(m_1-\alpha_1)(t_T)^{m_1 + k - \alpha_1} T^{j+1-k} }{(j+1-k)!\Gamma(m_1 + k + 1 - \alpha_1} \right] + \frac{(j+1)!\Gamma(m_1 - \alpha_1)t^{m_1+k-\alpha_1}}{\Gamma(m_1 + j + 1 - \alpha_1)} \right] + \frac{a^{\prime}_j}{j+1}\left[ \sum_{k=0}^{k=j+1} \frac{(j+1)!\Gamma(m_1 - \alpha_1)(t-T)^{m_1+k-\alpha_1}T^{j+1-k}}{(j+1-k)!\Gamma(m_1 + k + 1 -\alpha_1)} \right] \right] }{ \frac{1}{\Gamma(m_2-\alpha_2)} \frac{d^{m_2}}{dt^{m_2}} \sum_{j=0}^{j=5} \left[ \frac{b_j}{j+1} \left[ \sum_{k=0}^{k=j+1} \left[ \frac{-(j+1)!\Gamma(m_2 - \alpha_2)(t-T)^{m_2 + k - \alpha_2} T^{j+1-k}}{(j+1-k)!\Gamma(m_2 + k + 1 - \alpha_2)} \right] + \frac{(j+1)!\Gamma(m_2 - \alpha_2)t^{m_2 + k - \alpha_2}}{\Gamma(m_2 + j + 1 - \alpha_2)} \right] + \frac{b^{\prime}_j}{j+1}\left[ \sum_{k=0}^{k=j+1} \frac{(j+1)!\Gamma(m_2 - \alpha_2)(t-T)^{m_2 + k - \alpha_2}T^{j+1-k}}{(j+1-k)!\Gamma(m_2 + k + 1 - \alpha_2)} \right] \right]}\\
        & = \frac{ \frac{1}{\Gamma(m_1 - \alpha_1)} \frac{d^{m_1}}{dt^{m_1}} \sum_{j=0}^{j=5} \left[ (a^{\prime}_j - a_j) \sum_{k=0}^{k = j + 1} \left[ \frac{j!\Gamma(m_1 - \alpha_1)(t - T)^{m_1 + k - \alpha_1} T^{j + 1 - k}}{(j+1-k)!\Gamma(m_1 + k + 1 - \alpha_1)} \right] + a_j \frac{j! \Gamma(m_1 - \alpha_1)t^{m_1 + j + 1 - \alpha_1}}{\Gamma(m_1 + j + 2 - \alpha_1} \right]}{ \frac{1}{\Gamma(m_2 - \alpha_2)} \frac{d^{m_2}}{dt^{m_2}} \sum_{j=0}^{j=5} \left[ (b^{\prime}_j - b_j) \sum_{k=0}^{k=j+1} \left[ \frac{j!\Gamma(m_2 - \alpha_2)(t-T)^{m_2 + k - \alpha_2}T^{j+1-k}}{(j+1-k)!\Gamma(m_2 + k + 1 - \alpha_2)} \right] + b_j \frac{j!\Gamma(m_2 - \alpha_2) t^{m_2 + j + 1 - \alpha_2}}{\Gamma(m_2 + j + 2 - \alpha_2)} \right] }\\ 
        & = \frac{\sum_{j=0}^{j=5} \left[ (a^{\prime}_j - a_j) \sum_{k=0}^{k=j+1} \left[ \frac{j!(t-T)^{k-\alpha_1}T^{j+1-k}}{(j+1-k)!\Gamma(k+1-\alpha_1)} \right] + a_j \frac{j!t^{j+1-\alpha_1}}{\Gamma(j+2-\alpha_1)}\right] }{\sum_{j=0}^{j=5} \left[ (b^{\prime}_j - b_j) \sum_{k=0}^{k=j+1} \left[ \frac{j!(t-T)^{k - \alpha_2}T^{j+1-k}}{(j+1-k)!\Gamma(k+1-\alpha_2)} \right] + b_j \frac{j!t^{j+1-\alpha_2}}{\Gamma(j+2-\alpha_2)} \right]}
    \end{split}
\end{equation}
\end{landscape}

\added[id=RL]{In this 2-piece wise approximation, the vertex is non-differentiable, this implies that (16) expression has a singularity at T (because $(t-T)^{-\alpha_{1,2} }\rightarrow \infty)$.}

\added[id=RL]{It could be possible to avoid this singularity, using a 3-piece wise approximation, smoothing the vertex. However, the calculus are very tedious. We will explain, below, what our simpler choice implies.}

\begin{equation}
    \begin{split}
    \text{Then }F_M^{\alpha_1,\alpha_2}(t)& = \frac{(t-T)^{-\alpha_1}\left[ \sum_{j=0}^{j=5}\left[ (a^{\prime}_j - a_j) \sum_{k=0}^{k=j+1} \left[ \frac{j!(t-T)^kT^{j+1-k}}{(j+1-k)!\Gamma(k+1-\alpha_1}) \right]  + a_j \frac{j!t^{j+1 - \alpha_1}(t-T)^{\alpha_1}}{\Gamma(j+2-\alpha_1)} \right] \right] }{ (t-T)^{-\alpha_2} \left[ \sum_{j=0}^{j=5} \left[ (b^{\prime}_j - b_j) \sum_{k=0}^{k=j+1} \left[ \frac{j!(t-T)^k T^{j+1-k}}{(j+1-k)!\Gamma(k+1-\alpha_2)} \right] + b_j \frac{j!t^{j+1-\alpha_2}(t-T)^{\alpha_2}}{\Gamma(j+2-\alpha_2)} \right]  \right] }\\
    & = \frac{ \sum_{j=0}^{j=5} \left[ (a^{\prime}_j - a_j) \sum_{k=0}^{k=j+1} \left[ \frac{j!(t-T)^kT^{j+1-k}}{(j+1-k)!\Gamma(k+1-\alpha_1)} \right] + a_j\frac{j!t^{j+1-\alpha_1}(t-T)^{\alpha_1}}{\Gamma(j+2-\alpha_1)} \right] }{ (t-T)^{\alpha_1 - \alpha_2} \sum_{j=0}^{j=5}\left[ (b^{\prime}_j - b_j) \sum_{k=0}^{k=j+1} \left[ \frac{j!(t-T)^kT^{j+1-k}}{(j+1-k)!\Gamma(k+1-\alpha_2)} \right] + b_j\frac{j!t^{j+1-\alpha_2}(t-T)^{\alpha_2}}{\Gamma(j+2 - \alpha_2)} \right] }
    \end{split}
\end{equation}

\added[id=RL]{Finally}

\begin{equation}
    F_M^{\alpha_1,\alpha_2}(t) = \begin{cases}
    \frac{\sum_{j=0}^{j=5} \frac{a_j\Gamma(j+1)}{\Gamma(j+2-\alpha_1)}t^{j+1-\alpha_1} }{\sum_{j=0}^{j=5}\frac{b_j\Gamma(j+1)}{\Gamma(j+2-\alpha_2)}t^{j+1-\alpha_2}}\text{,} & \text{for} 0 \leq t \leq T \\
    \frac{\sum_{j=0}^{j=5} \left[ (a^{\prime}_j - a_j) \sum_{k=0}^{k=j+1} \left[ \frac{j!(t-T)^kT^{j+1-k}}{(j+1-k)!\Gamma(k+1-\alpha_1)} \right] + a_j \frac{j!t^{j+1-\alpha_1}(t-T)^{\alpha_1}}{\Gamma(j+2-\alpha_1)}\right] }{(t-T)^{\alpha_1-\alpha_2} \sum_{j=0}^{j=5} \left[ (b^{\prime}_j - b_j) \sum_{k=0}^{k=j+1} \left[ \frac{j!(t-T)^kT^{j+1-k}}{(j+1-k)!\Gamma(k+1-\alpha_2)} \right] + b_j\frac{j!t^{j+1-\alpha_2(t-T)^{\alpha_2}}}{\Gamma(j+2-\alpha_2)} \right] }\text{,} & \text{for} T < t < 171
    \end{cases}
\end{equation}

\begin{figure}[!tpb]
    \centering
    \includegraphics[width=0.8\textwidth]{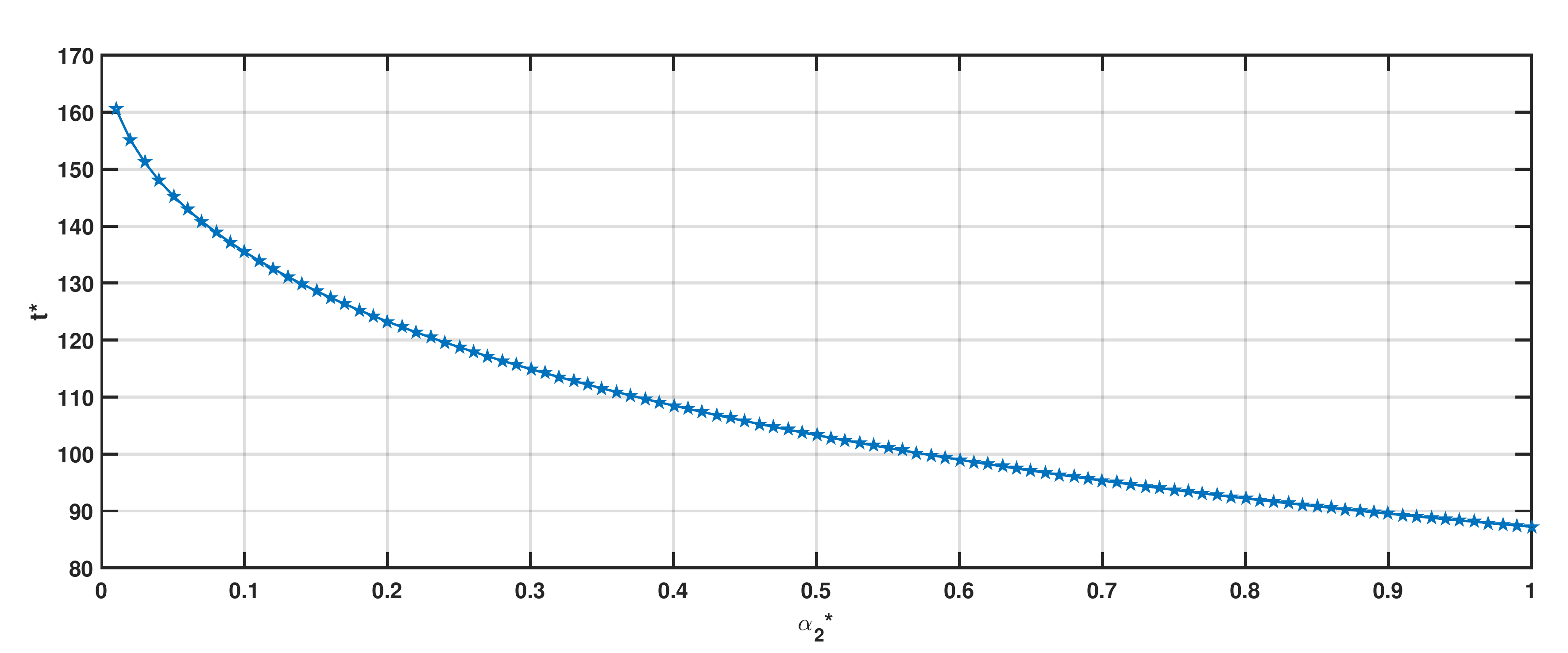}
    \caption{The first zero $t^*(\alpha_2)\geq T$, of the denominator of $F_M^{\alpha_1,\alpha_2}(t)$, as function of $\alpha_2$.}
    \label{fig:fig25}
\end{figure}

\added[id=RL]{Step 4 choice of parameter $\alpha_1$ and $\alpha_2$: Following the same idea as for the first alternative, we try to avoid singularity for $F_M^{\alpha_1,\alpha_2}(t)$, except of course the singularity near $T$, which is of mathematical nature (non-differentiability of voltage and intensity at $t = T$). Figure~\ref{fig:fig25} display the first zero $t^* (\alpha_2)\geq T$, of the denominator of $F_M^{\alpha_1,\alpha_2}(t)$. One can see that $t^*(1)\cong T$.}

\begin{figure}[!tpb]
    \centering
    \includegraphics[width=0.8\textwidth]{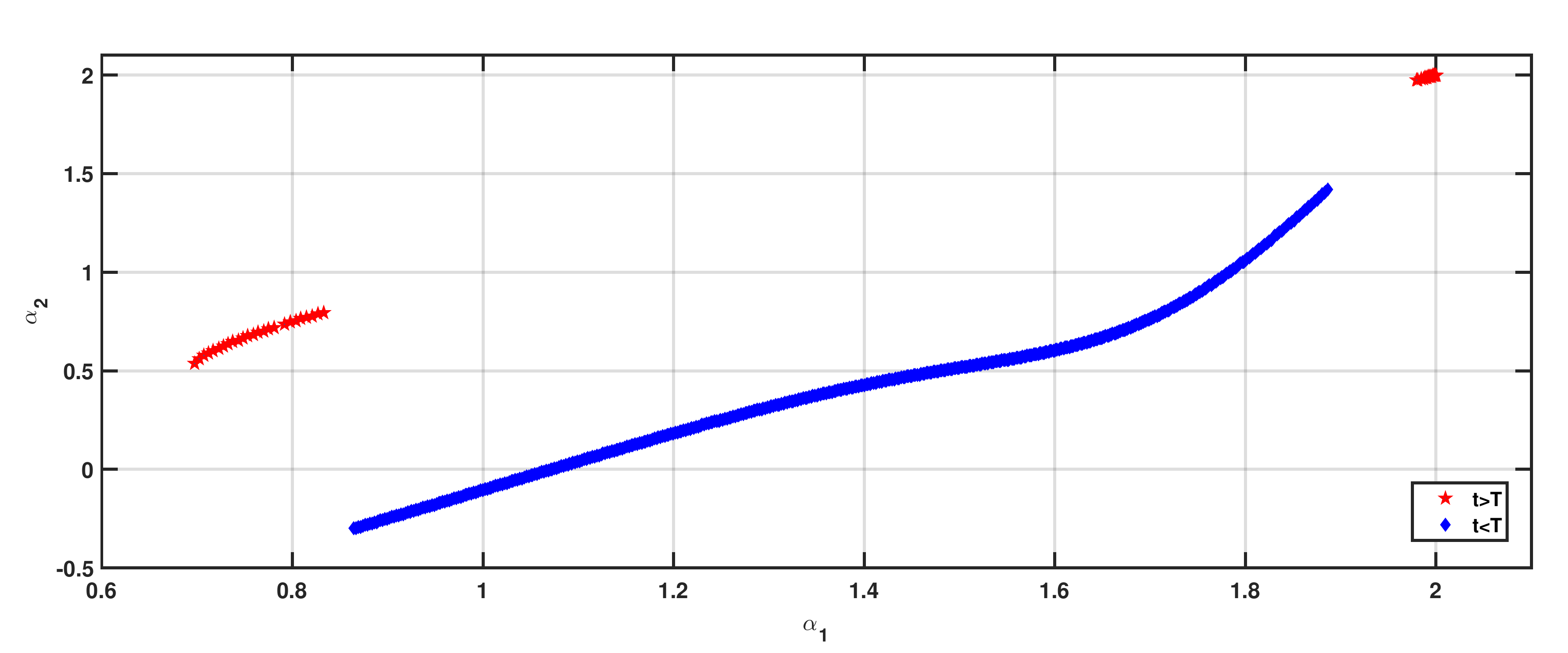}
    \caption{Couples $(\alpha_1,\alpha_2)$ for which the denominator and numerator of $F_M^{\alpha_1,\alpha_2}(t)$ are null simultaneously for $t < T$ (blue dot) and $t > T$ (red dot).}
    \label{fig:fig26}
\end{figure}

\begin{figure}[!tpb]
    \centering
    \includegraphics[width=0.8\textwidth]{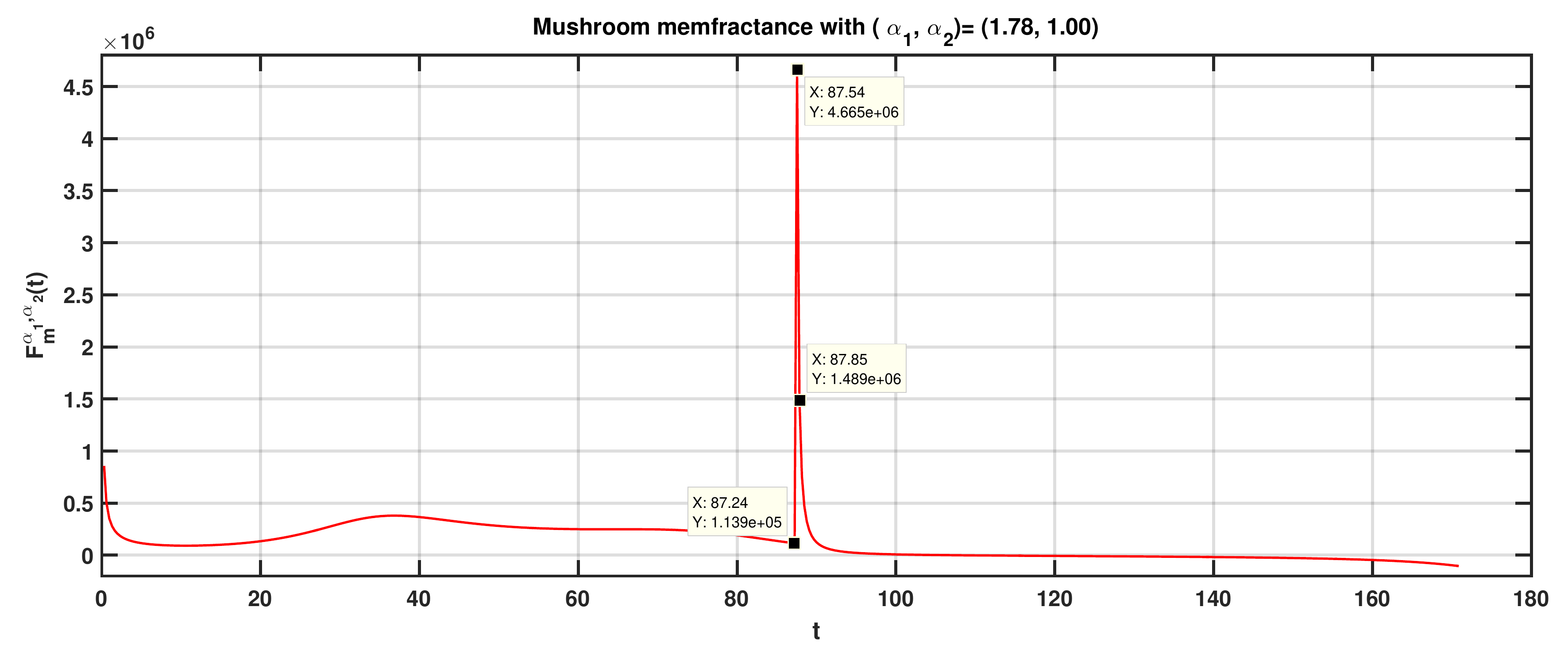}
    \caption{Memfractance for $(\alpha_1=1.78,\alpha_2=1.00)$ given in Table~\ref{tab:tab5}}
    \label{fig:fig27}
\end{figure}

\added[id=RL]{Figure~\ref{fig:fig26} displays the curves of couples $(\alpha_1,\alpha_2)$ for which the denominator and numerator of $F_M^{\alpha_1,\alpha_2}(t)$ are null simultaneously  for $t < T$ and $t > T$. On  this figure, the value of $\alpha_1$, that  corresponds to $\alpha_2=1$  is  $\alpha_1\approx1.78348389322388$. The corresponding Memfractance is displayed in Fig.~\ref{fig:fig27}.}

\begin{figure}[!tpb]
    \centering
    \includegraphics[width=0.8\textwidth]{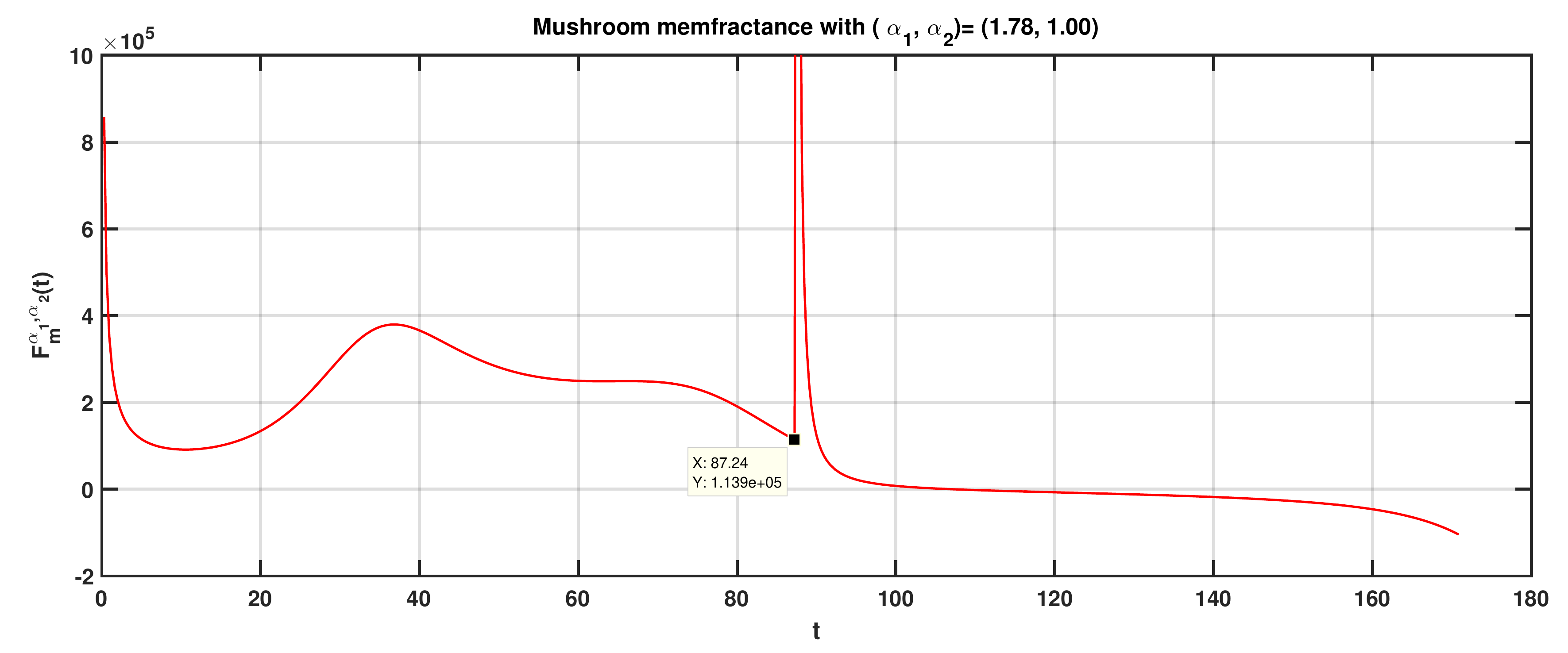}
    \caption{Magnification of Fig.~\ref{fig:fig27}}
    \label{fig:fig28}
\end{figure}

\added[id=RL]{The singularity observed in Figs.~\ref{fig:fig27}-\ref{fig:fig28} is due to the non-differentiability of both voltage and intensity functions at point $T$. It is only a mathematical problem of approximation which can be solved using a 3-piecewise polynomial instead of the 2-piecewise polynomial ($P1(t)$, $P2(t)$) and ($P3(t)$, $P4(t)$). The third added piecewise polynomials for $v(t)$ and $i(t)$ being defined on the tiny interval [87.24, 87.90]. However due to more tedious calculus, we do not consider this option in the present article. It is only a math problem, and one can consider that Fig. 28 represents the value of the memfractance in the interval $[0,87.24]\cup[87.90,171]$.}

\begin{figure}[!tpb]
    \centering
    \includegraphics[width=0.8\textwidth]{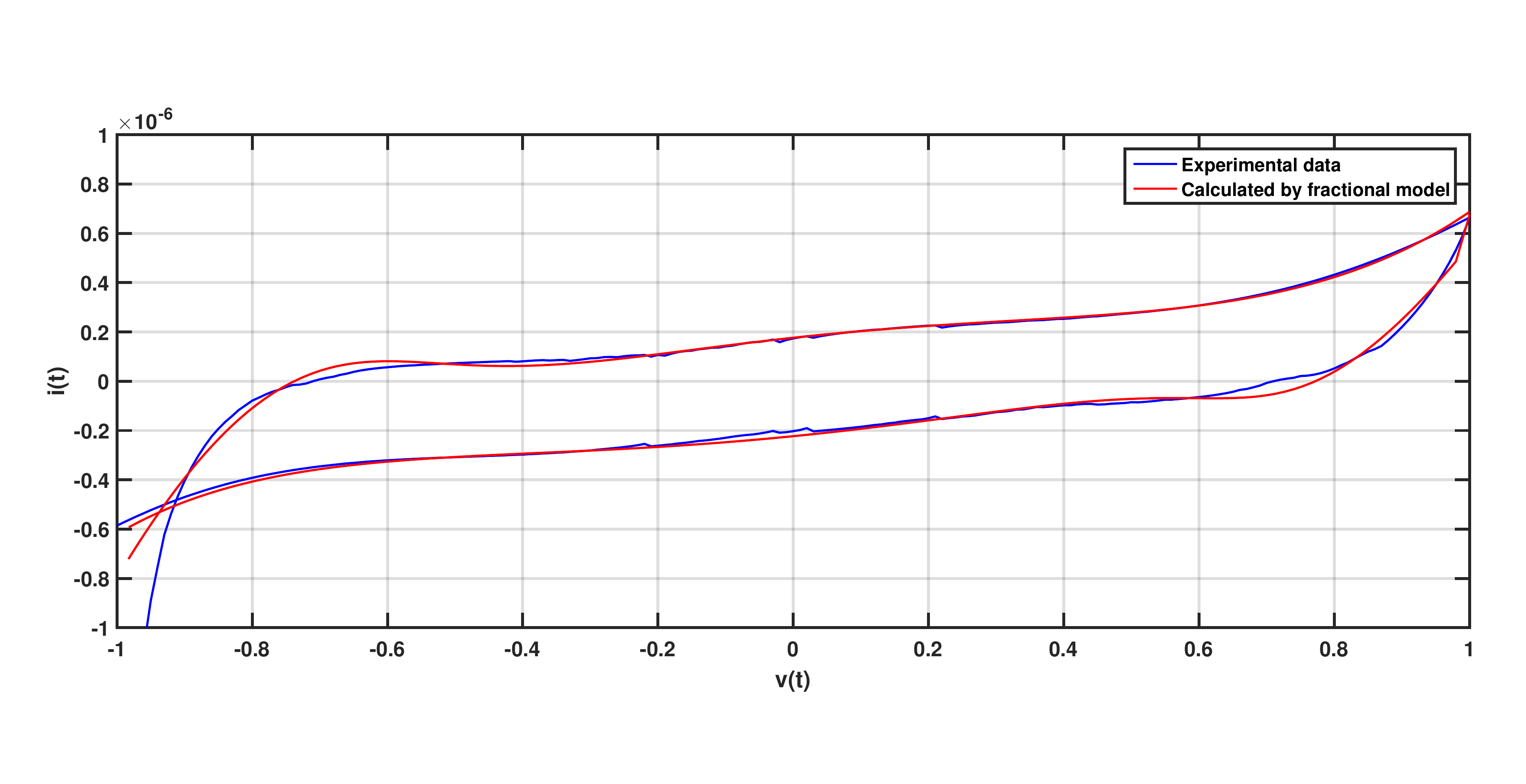}
    \caption{Comparison between average experimental data of cyclic voltammetry performed over -1\,V to 1\,V, Stem-to-cap electrode placement, and closed approximative formula.}
    \label{fig:fig29}
\end{figure}

\added[id=RL]{The value of $(\alpha_1=1.78,\alpha_2=1.00)$ belongs to the line segment of Fig.~\ref{fig:mem-fractor_space}, whose extremities are Memristor, and Capacitor. Which means that Oyster mushroom fruit bodies with stem to cap electrodes, is like a mix of such basic electronic devices.
The comparison of average experimental data of cyclic voltammetry performed over -1\,V to 1\,V, Stem-to-cap electrode placement, and closed approximative formula is displayed in Fig.~\ref{fig:fig29}, showing a very good agreement between both curves.
}

\section{Discussion}
\label{sec:discussions}

\added[id=RL]{
First at all, we have two remarks:

	\textbf{1.} Both approximations used in Section~\ref{sec:model} converge to Memfractance with parameter value $(\alpha_1,\alpha_2)$ - belonging inside or on edge of the triangle of Fig.~\ref{fig:mem-fractor_space}, whose vertices are Memristor, Memcapacitor and Capacitor. 
	
Of course, the value for these approximations are not exactly the same. This is due, in part to the fact that we consider that the most regular approximation is the one for which the function range $(F_M^{\alpha_1,\alpha_2}(t))$ is minimal. Other choices based on physiology of Mushroom could be invoked.
Moreover, the Memfractance is computed on the averaged curve of 20 runs which do not present exactly the same characteristic voltammetry. Oyster mushroom fruit bodies are biological material, which prevent exact reproduction of electrical property.

	\textbf{2.} The use of fractional derivatives to analyze the memfractance, is obvious if one considers that fractional derivatives have memory, which allow a perfect modeling of memristive elements. Their handling is however delicate if one wants to avoid any flaw.

}

Similar I-V characteristics have been experienced for slime mould~\cite{gale2015slime} and apples~\cite{paper:apples_memristor}. The cyclic voltammetry experiments demonstrate that the I-V curve produced from these living substrates is a closed loop where the negative path does not match the positive path. Hence the fungi display the characteristics of a memristor . A similar conclusion is drawn for the microtubule experiments~\cite{paper:tublin_microtubule_droplets}. The microtubule exhibits different resistive properties for the same applied voltage depending on the history of applied voltages.\par 

Additionally, the fruit bodies produce current oscillations during the cyclic voltammetry. This oscillatory effect is only observed on one phase of the voltammetry for a given voltage range which is, again, a behaviour that can be associated to a device whose resistance is a function of its previous resistance. This spiking activity is typical of a device that exhibits memristive behaviours. Firstly, it was reported in experiments with electrochemical devices using graphite reference electrodes, that a temporal dependence of the current of the device - at constant applied voltage - causes charge accumulation and discharge~\cite{erokhin2008electrochemically}. The spiking is also apparent in some plots, for a large electrode size, in experiments with electrode metal on solution-processed flexible titanium dioxide memristors~\cite{gale2015effect}. A detailed analysis of types of spiking emerging in simulated memristive networks was undertaken in~\cite{gale2014emergent}. Repeatable observations of the spiking behaviour in I-V of the fungi is very important because this opens new pathways for the implementation of neuromorphic computing with fungi. A fruitful theoretical foundation of this field is already well developed~\cite{serrano2013proposal,indiveri2013integration,prezioso2016spiking,pickett2013scalable,linares2011spike,indiveri2015memory}.

\section{Conclusion}
\label{sec:conclusions}

The fruit bodies of grey oyster fungi \emph{Pleurotus ostreatus} were subjected to I-V characterisation a number of times, from which it was clearly shown that they exhibit mem\replaced[id=AB]{-fractor}{ristive} properties. Under cyclic voltammetry, the fruit body will conduct differently depending on the phase (positive or negative) of the voltammetry. This behaviour produces the classic ``lobes''  in the I-V characteristics of a memristor. \par

However, a biological medium, such as the fruit body of the grey oyster fungi presented here, will differ from that of the ideal memristor model since the ``pinching'' behaviour and size of the hysteresis lobes are functions of the frequency of the voltage sweep as well as the previous resistance. Typically, the biological medium generates its own potential across the electrodes, therefore, even when no additional potential is supplied, there is still current flow between the probes. \added[id=AB]{This property of the fungi produces an opening in the I-V curve that is a classic property of the mem-capacitor. Since the fungi are exhibiting properties of both memristors and mem-capacitors, their electrical memory behaviour puts them somewhere in the mem-fractor solution space where $ 0 < \alpha_1, \alpha_2 < 1$. Hence, it has been shown that fungi act as mem-fractors}.\par

\section*{Acknowledgement}

This project has received funding from the European Union's Horizon 2020 research and innovation programme FET OPEN ``Challenging current thinking'' under grant agreement No 858132.

\section*{Author contributions}
A.A. conceived the idea of experiments. A.A. and A.P. prepared the substrate colonised by mycelium. A.B. performed experiments, collected data and produced all plots in the manuscript.
R.L. and MS.A. derived the mathematical model of mem-fractance as seen in grey oyster fungi.
All authors prepared manuscript (wrote and reviewed all contents).

\bibliographystyle{plain}
\bibliography{references, references_model}
\end{document}